\documentclass[10pt,twocolumn,letterpaper]{article}

\usepackage{iccv}
\usepackage{times}
\usepackage{epsfig}
\usepackage{graphicx}
\usepackage{amsmath}
\usepackage{amssymb}

\usepackage{xfrac}
\usepackage{subcaption}
\usepackage[export]{adjustbox}
\usepackage{rotating}
\usepackage{array}
\usepackage{color}
\usepackage{booktabs}
\usepackage{tablefootnote}
\usepackage{colortbl}
\usepackage{tabularx}
\usepackage{multirow}

\usepackage[nolist]{acronym}
\usepackage{units}

\usepackage{pgfplots}
\usepackage{tikz} 
\pgfplotsset{compat=newest}
\usepgfplotslibrary{groupplots}
\usetikzlibrary{matrix}

\usepackage[pagebackref=true,breaklinks=true,letterpaper=true,colorlinks,bookmarks=false]{hyperref}

\iccvfinalcopy % *** Uncomment this line for the final submission

\def\iccvPaperID{3718} % *** Enter the ICCV Paper ID here

\ificcvfinal\fi
\begin{document}

%%%%%%%%% TITLE
\title{\vspace{-5pt}Gated2Depth: Real-Time Dense Lidar From Gated Images}

\author{
	Tobias Gruber$^{1,3}$\hspace{0.12in}
	Frank Julca-Aguilar$^{2}$\hspace{0.12in}
	Mario Bijelic$^{1,3}$\hspace{0.12in}
       Felix Heide$^{2,4}$\hspace{0.12in} \vspace{8pt}
	\\ 
	\textsuperscript{1}Daimler AG\hspace{0.15in}
	\textsuperscript{2}Algolux\hspace{0.15in}
	\textsuperscript{3}Ulm University\hspace{0.15in}
	\textsuperscript{4}Princeton University\hspace{0.15in} 
	\vspace{-5pt}
}

\maketitle

\ificcvfinal\thispagestyle{empty}\fi

% formatting stuff

\definecolor{Gray}{rgb}{0.5,0.5,0.5}
\definecolor{darkblue}{rgb}{0,0,0.7}
\definecolor{orange}{rgb}{1,.5,0} % something readable but different from todo
\definecolor{red}{rgb}{1,0,0} % something readable but different from todo

% taken from https://designnavigator.daimler.com/Daimler_Color_System
\definecolor{dai_ligth_grey}{RGB}{230,230,230}
\definecolor{dai_ligth_grey20K}{RGB}{200,200,200}
\definecolor{dai_ligth_grey40K}{RGB}{158,158,158}
\definecolor{dai_ligth_grey60K}{RGB}{112,112,112}
\definecolor{dai_ligth_grey80K}{RGB}{68,68,68}
\definecolor{dai_petrol}{RGB}{0,103,127}
\definecolor{dai_petrol20K}{RGB}{0,86,106}
\definecolor{dai_petrol40K}{RGB}{0,67,85}
\definecolor{dai_petrol80}{RGB}{0,122,147}
\definecolor{dai_petrol60}{RGB}{80,151,171}
\definecolor{dai_petrol40}{RGB}{121,174,191}
\definecolor{dai_petrol20}{RGB}{166,202,216}
\definecolor{dai_deepred}{RGB}{113,24,12}
\definecolor{dai_deepred20K}{RGB}{90,19,10}
\definecolor{dai_deepred40K}{RGB}{68,14,7}
%\definecolor{violettblau}{cmyk}{0.9, 0.6, 0, 0}
\definecolor{rot}{RGB}{238, 28 35}
\definecolor{apfelgruen}{RGB}{140, 198, 62}
%\definecolor{gelb}{RGB}{255, 229, 0}
\definecolor{orange}{RGB}{244, 111, 33}
\definecolor{pink}{RGB}{237, 0, 140}
\definecolor{lila}{RGB}{128, 10, 145}
%\definecolor{hellgrau}{RGB}{224, 224, 224}
%\definecolor{mittelgrau}{RGB}{128, 128, 128}
%\definecolor{dunkelgrau}{RGB}{80,80,80}
\definecolor{anthrazit}{RGB}{19, 31, 31}

\newcommand{\heading}[1]{\noindent\textbf{#1}}
\newcommand{\note}[1]{{\em{\textcolor{orange}{#1}}}}
\newcommand{\todo}[1]{{\textcolor{red}{\bf{TODO: #1}}}}
\newcommand{\comments}[1]{{\em{\textcolor{orange}{#1}}}}
\newcommand{\changed}[1]{#1}
\newcommand{\place}[1]{ \begin{itemize}\item\textcolor{darkblue}{#1}\end{itemize}}
\newcommand{\de}{\mathrm{d}}

\newcommand{\normlzd}[1]{{#1}^{\textrm{aligned}}}

% dimensions
\newcommand{\ttime}{\tau}               % time coordinate
\newcommand{\x}{\Vect{x}}               % spatial coordinates in vectorized form
\newcommand{\z}{z}               % depth coordinate of volume

\newcommand{\npixels}{n}               % num pixels
\newcommand{\ntime}{t}               % num timesteps

%image formation
\newcommand{\illfunc}     {g}
\newcommand{\pathfunc}     {s}
\newcommand{\camfunc}     {f}

% notation for image formation
\newcommand{\irradiance}{E}
\newcommand{\exposure}{b}
\newcommand{\pmdfunc}{f}                % modulation on the PMD side
\newcommand{\lightfunc}{g}              % modulation of the light
\newcommand{\period}{T}                 % temporal period of the modulation
\newcommand{\freqm}{\omega}                % frequency of modulation
\newcommand{\illphase}{\rho}             % frequency of modulation harmonics
\newcommand{\sensphase}{\psi}             % frequency of modulation harmonics of pmd camera
\newcommand{\pmdphase}{\phi}            % phase of PMD modulation
\newcommand{\omphi}{{\omega,\phi}}      % shorthand for freq/phase pair
\newcommand{\numperiod}{N}              % #periods integrated over
\newcommand{\att}{\alpha}               % geometric & photometric attenuation
\newcommand{\pathspace}{{\mathcal{P}}}  % space of all light paths

% regular modulation based cameras
\newcommand{\atan}{\operatorname{atan}}

% math stuff
\newcommand{\Fourier}{\mathfrak{{F}}}         % fourier transform
\newcommand{\conv}     {\otimes}
\newcommand{\corr}     {\star}
\newcommand{\Mat}[1]    {{\ensuremath{\mathbf{\uppercase{#1}}}}} %Matrix 
\newcommand{\Vect}[1]   {{\ensuremath{\mathbf{\lowercase{#1}}}}} %Vector
\newcommand{\Id}				{\mathbb{I}} %Identity matrix
\newcommand{\Diag}[1] 	{\operatorname{diag}\left({ #1 }\right)} %Diagonalized matrix
\newcommand{\Opt}[1] 	  {{#1}_{\text{opt}}} %Optimal point of an optimization
\newcommand{\CC}[1]			{{#1}^{*}} %Convex conjugate
\newcommand{\Op}[1]     {\Mat{#1}} %Operator
\newcommand{\minimize}[1] {\underset{{#1}}{\operatorname{argmin}} \: \: } %Minimize w.r.t.
\newcommand{\maximize}[1] {\underset{{#1}}{\operatorname{argmax}} \: \: } %Maximize w.r.t.  
\newcommand{\grad}      {\nabla}

% notation for method
\newcommand{\Basis}{\Mat{H}}         		% Matrix basis
\newcommand{\Corr}{\Mat{C}}             % measurement matrix
\newcommand{\correlem}{\bold{c}}             % measurement matrix element
\newcommand{\meas}{\Vect{b}}            % measurement vector
\newcommand{\Meas}{\Mat{B}}            % measurement matrix
\newcommand{\MeasNormalized}{\Mat{B}^{\textrm{new}}}            % measurement matrix
\newcommand{\Img}{H}                    % transient image
\newcommand{\img}{\Vect{h}}             % vectorized image
\newcommand{\latentresponse}{\alpha}

\newenvironment{customlegend}[1][]{%
        \begingroup
        % inits/clears the lists (which might be populated from previous
        % axes):
        \csname pgfplots@init@cleared@structures\endcsname
        \pgfplotsset{#1}%
    }{%
        % draws the legend:
        \csname pgfplots@createlegend\endcsname
        \endgroup
    }%

    % makes \addlegendimage available (typically only available within an
    % axis environment):
    \def\addlegendimage{\csname pgfplots@addlegendimage\endcsname}

%%%%%%%%% ABSTRACT
\begin{abstract}
\vspace{-6pt}
   We present an imaging framework which converts three images from a gated camera into high-resolution depth maps with depth accuracy comparable to pulsed lidar measurements. Existing scanning lidar systems achieve low spatial resolution at large ranges due to mechanically-limited angular sampling rates, restricting scene understanding tasks to close-range clusters with dense sampling. Moreover, today's pulsed lidar scanners suffer from high cost, power consumption, large form-factors, and they fail in the presence of strong backscatter. We depart from point scanning and demonstrate that it is possible to turn a low-cost CMOS gated imager into a dense depth camera with at least \unit[80]{m} range -- by learning depth from three gated images. The proposed architecture exploits semantic context across gated slices, and is trained on a synthetic discriminator loss without the need of dense depth labels. The proposed replacement for scanning lidar systems is real-time, handles back-scatter and provides dense depth at long ranges. We validate our approach in simulation and on real-world data acquired over 4,000~km driving in northern Europe. Data and code are available at \small{\url{https://github.com/gruberto/Gated2Depth}}.
\end{abstract}
	
%%%%%%%%%% BODY TEXT
\vspace*{-4mm}
\section{Introduction}
\begin{figure*}[t!]
	\centering
	\includegraphics[width=0.95\textwidth]{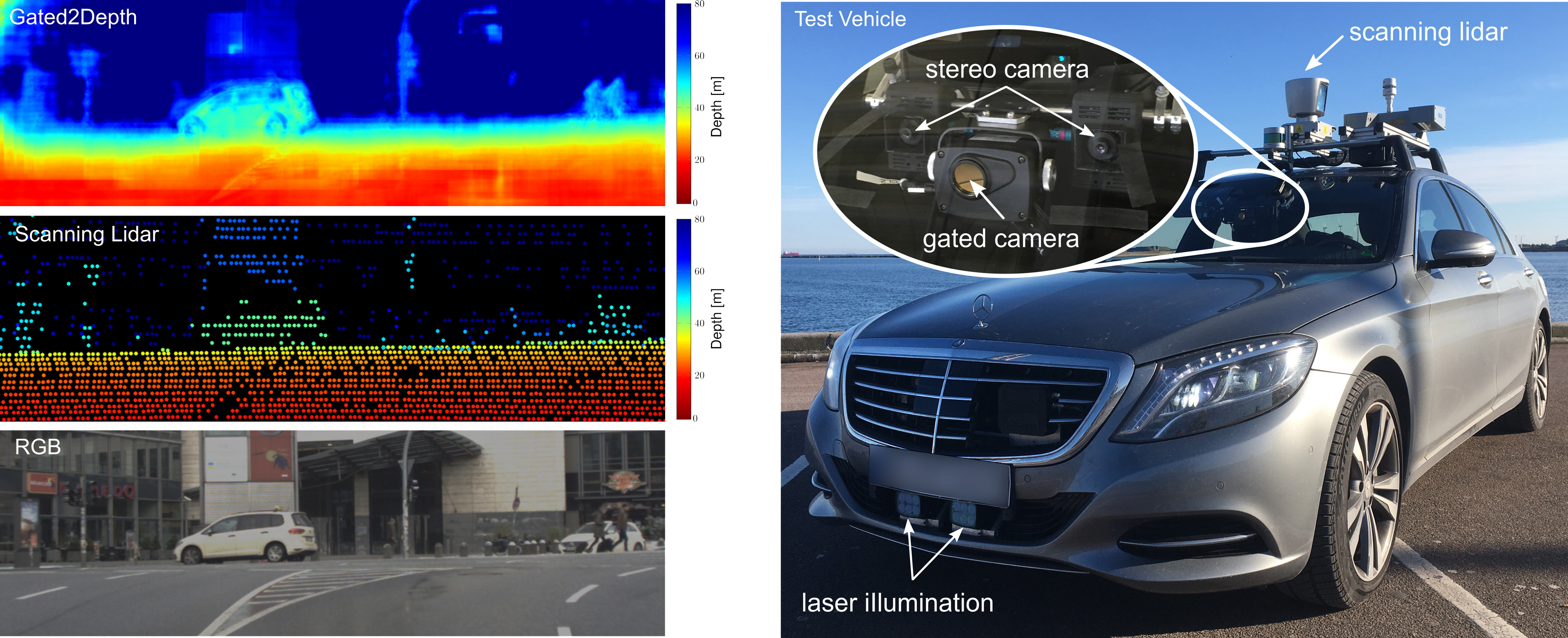}
	\caption{We propose a novel real-time framework for dense depth estimation (top-left)
		without scanning mechanisms. Our method maps measurements from a flood-illuminated gated camera behind the wind-shield (inset right), captured in real-time, to dense depth maps with depth accuracy comparable to lidar measurements (center-left). In contrast to the sparse lidar measurements, these depth maps are high-resolution enabling semantic understanding at long ranges. We evaluate our method on synthetic and real data, collected with a testing and a scanning lidar Velodyne HDL64-S3D as reference (right).
	}
	\label{fig:Teaser}
	\vspace*{-10pt}
\end{figure*}

Active depth cameras, such as scanning lidar systems, have not only become a cornerstone imaging modality for autonomous driving and robotics, but are emerging in applications across disciplines, including autonomous drones, remote sensing, human-computer interaction, and augmented or virtual reality. Depth cameras that provide dense range allow for dense scene reconstructions~\cite{izadi2011kinectfusion} when combined with color cameras, including correlation time-of-flight cameras (C-ToF)~\cite{hansard2012time,kolb2010time,lange00tof} such as Microsoft's Kinect One, or structured light cameras~\cite{achar2017epipolar,otoole2014temporal,o2012primal,scharstein2003high}. These acquisition systems facilitate the collection of large-scale RGB-D data sets that fuel research on core computer vision problems, including scene understanding~\cite{hickson2014efficient,song2015sun} and action recognition~\cite{ni2013rgbd}. However, while existing depth cameras provide high-fidelity depth for close ranges indoors~\cite{izadi2011kinectfusion,Newcombe_2015_CVPR}, dense depth imaging at long ranges and in dynamic outdoor scenes is an open challenge.

\begin{figure}[t!]
	\setlength\tabcolsep{1.5pt}
	\centering
	\begin{tabular}{ccc}
	RGB Camera & Gated Camera & Lidar Bird's Eye View \\
	\includegraphics[height=0.145\columnwidth]{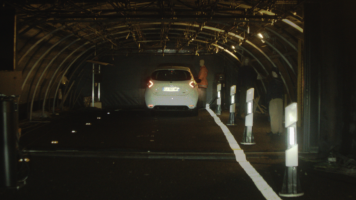} &
	\includegraphics[height=0.145\columnwidth]{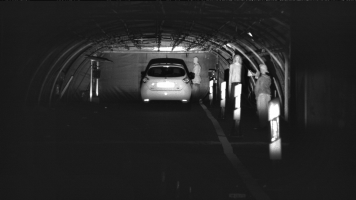} & 
	\includegraphics[height=0.145\columnwidth]{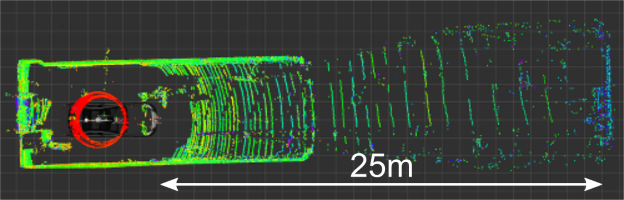} \\[-1pt]
	\includegraphics[height=0.145\columnwidth]{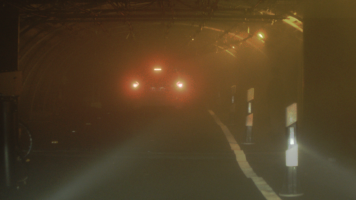} &
	\includegraphics[height=0.145\columnwidth]{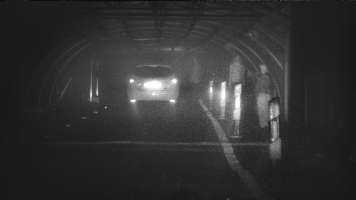} & 
	\includegraphics[height=0.145\columnwidth]{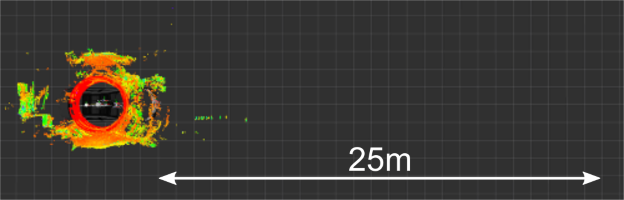} \\
	\end{tabular}
	\vspace*{-10pt}
	\caption{Sensor performance in a fog chamber with very dense fog. The first row shows recordings without fog while the second row shows the same scene in dense fog.}
	\label{fig:sensor_performance}
	\vspace*{-5pt}
\end{figure}

Active imaging at long ranges is challenging because diffuse scene points only return a small fraction of the emitted photons back to the sensor. For perfect Lambertian surfaces, this fraction decreases quadratically with distance, posing a fundamental limitation as illumination power can only be increased up to the critical eye-safety level~\cite{schwarz2010lidar,spinhirne1995compact,williams_apds}. To tackle this constraint, existing pulsed lidar systems employ sensitive silicon avalanche photo-diodes (APDs) with high photon detection efficiency in the NIR band~\cite{williams_apds}. The custom semiconductor process for these sensitive detectors restricts current lidar systems to a single (or few) APDs instead of monolithic sensor arrays, which requires point-by-point scanning. Although scanning lidar approaches facilitate depth imaging at large ranges, scanning reduces their spatial resolution quadratically with distance, prohibiting semantic tasks for far objects, as shown in Figure~\ref{fig:Teaser}. 
Recently, single-photon avalance diodes (SPADs)~\cite{aull2002geiger,bronzi2016automotive,niclass2005design,rochas2003first} are emerging as a promising technology that may enable sensor arrays in the CMOS process~\cite{villa2014cmos} in the future. Although SPADs are sensitive to individual photons, existing designs are highly photon-inefficient due to very low fill factors around 1\%~\cite{veerappan2011160} and pile-up distortions at higher pulse powers~\cite{coates:68}. Moreover, passive depth estimation techniques do not offer a solution, including stereo cameras~\cite{hartley2003multiple,scharstein2003high} and depth from monocular imagery~\cite{eigen2014depth,Godard2017,saxena2006learning}. These approaches perform poorly at large ranges for small disparities, and they fail in critical outdoor scenarios, when ambient light is not sufficient, e.g. at night, and in the presence of strong back-scatter, e.g. in fog or snow, see Figure~\ref{fig:sensor_performance}.

Gated imaging is an emerging sensing technology that tackles these challenges by sending out pulsed illumination and integrating a scene response between temporal gates. Coarse temporal slicing allows for the removal of back-scatter due to fog, rain and snow, and can be realized in readily available CMOS technology. In contrast to pulsed lidar, gated imaging offers high signal measurements at long distances by integrating incoming photons over a large temporal slice, instead of time-tagging the first returns of individual pulses. However, although gated cameras offer an elegant, low-cost solution to outdoor imaging challenges, the sequential acquisition of the individual slices prohibits their use as depth cameras today, restricting depth information to a sparse set of wide temporal bins spanning more than \unit[50]{m} in depth. Note that using narrow slices does not offer a solution, because the slice width is inversely proportional to the number of captures, and thus frame-rate and narrow slices also means integrating less photons. With maximum frame rates of 120~Hz to 240~Hz, existing systems~\cite{grauer2014active} are limited to a range of 4 to 7 slices for dynamic scenes.

In this work, we present a method that recovers high-fidelity dense depth from sparse gated images. By learning to exploit semantic context across gated slices, the proposed architecture achieves depth accuracy comparable to scanning based lidar in large-range outdoor scenarios, essentially turning a gated camera into a low-cost dense flash lidar that captures dense depth at long distances and also sees through fog, snow and rain. The method jointly solves depth estimation, denoising, inpainting of missing or unreliable measurements, shadow and multi-path removal, while being highly efficient with real-time frame rates on consumer GPUs.

Specifically, we make the following contributions:
\begin{itemize}
	\setlength\itemsep{.2em}
	\item We introduce an image formation model and analytic depth estimation method using less than a handful of gated images.
	\item We propose a learning-based approach for estimating dense depth from gated images, without the need for dense depth labels for training. 
	\item We validate the proposed method in simulation and on real-world measurements acquired with a prototype system in challenging automotive scenarios. We show that the method recovers dense depth up to \unit[80]{m} with depth accuracy comparable to scanning lidar. 
	\item We provide the first long-range gated data set, covering over 4,000~km driving throughout northern Europe. The data set includes driving scenes in snow, rain, urban driving and sub-urban driving. 
\end{itemize}

%------------------------------------------------------------------------
\section{Related Work}
%-------------------------------------------------------------------------
\vspace{0.5em}\noindent\textbf{Depth Estimation from Intensity Images.}
A large body of work explores methods for extracting depth from conventional color image sensors. A first line of research on structure from motion methods sequentially captures a stack of monocular images and extracts geometry by exploiting temporal correlation in the stack~\cite{koenderink1991affine,torr1999feature,Ummenhofer2017, Zhou2017}. In contrast, multi-view depth estimation methods~\cite{hartley2003multiple} do not rely on sequential acquisition but exploit the disparity in simultaneously acquired image pairs~\cite{seitz2006comparison}. Recent approaches to estimating stereo correspondences allow for interactive frame-rates~\cite{Chang2018,Kendall2017,pilzer2018unsupervised}. Over the last years, a promising direction of research aims at estimating depth from a single monocular image~\cite{Chen2018b,eigen2014depth,Godard2017,Laina2016,saxena2006learning}, no longer requiring multi-view or sequential captures. 
Saxena et al.~\cite{saxena2006learning} introduce a Markov Random Field that incorporates multiscale image features for depth estimation. Eigen et. al~\cite{eigen2014depth} demonstrate that CNNs are well-suited for monocular depth estimation by learning priors on semantic-dependent depth~\cite{Chen2016,Godard2017,Laina2016}.
While consumer time-of-flight cameras facilitate the acquisition of large datasets for small indoor scenes~\cite{hickson2014efficient,song2015sun}, supervised training in large outdoor environments is an open challenge. Recent approaches tackle the lack of dense training data by proposing semi-supervised methods relying on relative depth \cite{Chen2016}, stereo images \cite{Garg2016,Godard2017,Kuznietsov2017}, sparse lidar points \cite{Kuznietsov2017} or semantic labels \cite{Ramirez2018}.
Passive methods have in common that their precision is more than an order of magnitude below that of scanning lidar systems which makes them no valid alternative to ubitious lidar ranging in autonomous vehicles~\cite{schwarz2010lidar}. In this work, we propose a method that allows to close this precision gap using low-cost gated imagers. 

%-------------------------------------------------------------------------
\vspace{0.5em}\noindent\textbf{Sparse Depth Completion.}
As an alternative approach to recover accurate dense depth, a recent work proposes depth completion from sparse lidar measurements. Similar to monocular depth estimation, learned encoder-decoder architectures have been proposed for this task \cite{chen2018estimating,jaritz2018sparse,ma2018sparse}. Jaritz et al.~\cite{jaritz2018sparse} propose to incorporate color RGB data for upsampling sparse depth samples but also require sparse depth samples in down-stream scene understanding tasks. To allow for an independent design of depth estimation and scene analysis algorithms, the completion architecture has to be trained with varying sparsity patterns \cite{jaritz2018sparse,ma2018sparse} or additional validity maps \cite{chen2018estimating}. While these depth completion methods offer improved depth estimates, they suffer from the same limitation as scanned lidar: low spatial resolution at long ranges due to limited angular sampling, low-resolution detectors, and costly mechanical scanning. 

%-------------------------------------------------------------------------
\vspace{0.5em}\noindent\textbf{Time-of-Flight Depth Cameras.}
Amplitude-modulated {C-ToF} cameras~\cite{hansard2012time,kolb2010time,lange00tof}, such as Microsoft's Kinect One, have become broadly adopted for indoor sensing~\cite{hickson2014efficient,song2015sun}. These cameras measure depth by recording the phase shift of periodically-modulated flood light illumination, which allows to extract the time-of-flight for the reflected flood light from the source to scene and back to the camera. However, in addition to the modulated light, this sensing approach also records all ambient background light. While per-pixel lock-in amplification removes background components efficiently in indoor scenarios~\cite{lange00tof}, and learned architectures can alleviate multi-path distortions~\cite{su2018deep}, existing C-ToF cameras are limited to ranges of a few meters in outdoor scenarios~\cite{heide2015doppler} in strong sunlight. 

Gated cameras send out pulses of flood-light and only record photons from a certain distance by opening and closing the camera after a given delay. Gated imaging has first been proposed by Heckman et al.~\cite{Heckman1967}. This acquisition mode allows to gate out backscatter from fog, rain, and snow~\cite{grauer2014active}. Busck et al.~\cite{Andersson2006,Busck2005,Busck2004} use gated imaging for high-resolution depth sensing by capturing large sequences of narrow gated slices. However, as the depth accuracy is inversely related to the gate width, and hence the number of required captures, sequentially capturing high-resolution gated depth is infeasible at real-time frame-rates. Recently, a line of research proposes analytic reconstruction models for known pulse and integration shapes~\cite{Laurenzis2009,Laurenzis2007,Xinwei2013}. These approaches require perfect knowledge of the integration and pulse profiles, which is impractical due to drift, and they provide low precision for broad gating windows in real-time capture settings. Adam et al.~\cite{adam2017bayesian}, and Schober et al.~\cite{schober2017dynamic}, present Bayesian methods for pulsed time-of-flight imaging of room-sized scenes. These methods solve probabilistic per-pixel estimation problems using priors on depth, reflectivity and ambient light, which is possible when using nanosecond exposure profiles~\cite{adam2017bayesian,schober2017dynamic} for room-sized scenes. In this work, we demonstrate that exploiting spatio-temporal scene semantics allows to recover dense and lidar-accurate depth from only three slices, with exposures two orders of magnitude longer ($>$ 100~ns), acquired in real-time. Using such wide exposure gates allows us to rely on low-cost gated CMOS imagers instead of detectors with high temporal resolution, such as SPADs.

%------------------------------------------------------------------------
\section{Gated Imaging}
\begin{figure}[t!]
    \centering
	\includegraphics[width=\columnwidth]{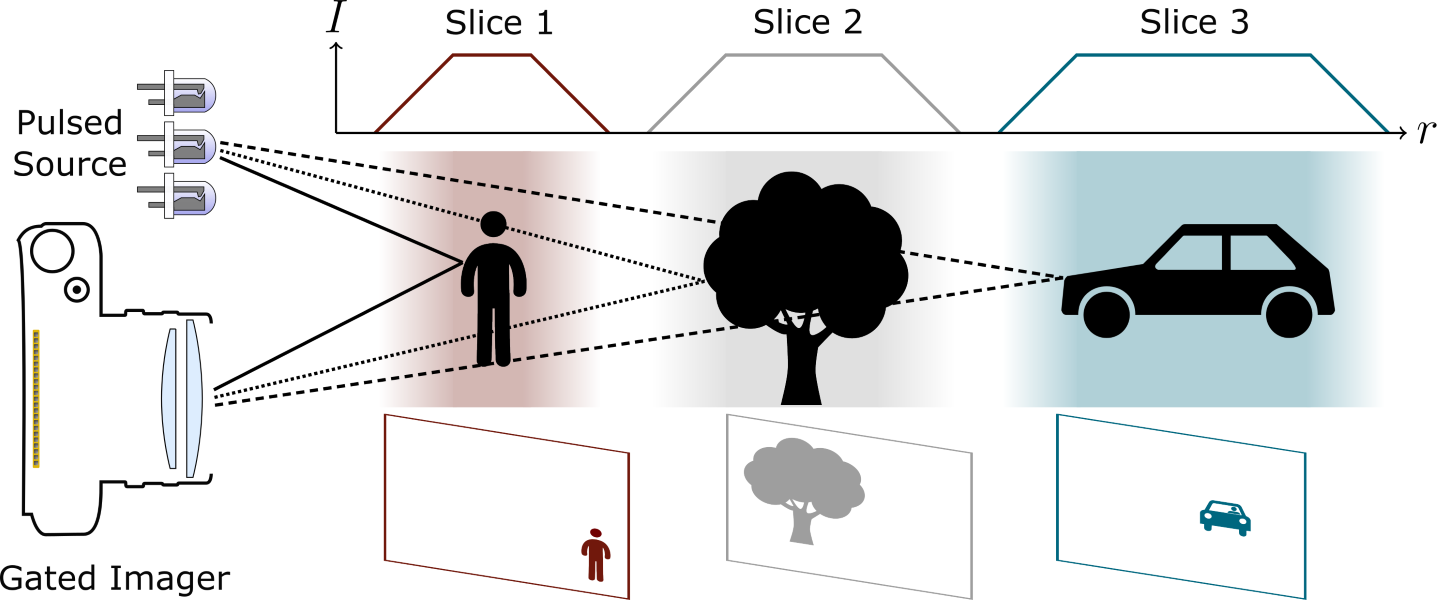}
    \vspace*{-15pt}
    \caption{A gated system consists of a pulsed laser source and a gated imager that are time synchronized. By setting the delay between illumination and image acquisition, the environment can be sliced into single images that contain only a certain distance range.}
    \label{fig:image_formation}
    \vspace*{-15pt}
\end{figure}

In this section, we review gated imaging and propose an analytic per-pixel depth estimation method. 
\vspace{-6pt}
\paragraph{Gated Imaging}\label{sec:gated_imaging}
Consider the setup shown in Figure~\ref{fig:image_formation}, where an amplitude-modulated source flood-illuminates the scene with broad rect-shaped ``pulses'' of light. The synchronized camera opens after a delay $\xi$ to receive only photons with round-trip path-length longer than $\xi \cdot c$, where $c$ is the speed of light. Assuming a dominating lambertian reflector at distance $r$, the detector gain is temporally modulated with the gating function $g$ resulting in the exposure measurement
\vspace{-7pt}
\begin{align}
I\left( r \right) = \alpha \; C\left( r \right) = \int\limits_{-\infty}^{\infty} g\left( t - \xi \right) \kappa\left( t, r \right) \textrm{d}t, \label{eq:gating_equation}
\vspace{-3pt}
\end{align}
where $\kappa$ is the temporal scene response, $\alpha$ the albedo of the reflector, and $C\left( r \right)$ the range-intensity profile. With the reflector at distance $r$, the temporal scene response can be described as
\vspace{-7pt}
\begin{align}\label{eq:scene_response}
\kappa \left( t, r \right) = \alpha p\left(t - \frac{2r}{c}\right) \beta\left( r \right).
\vspace{-3pt}
\end{align}
where $p$ is here the laser pulse profile and atmospheric effects, e.g. in a scattering medium, are modeled by the distance-dependent function $\beta$. Note that we ignore ambient light in Eq.~\eqref{eq:scene_response} which is minimized by a notch-filter in our setup and eliminated by subtraction with a separate capture without active illumination. In order to prevent the laser from overheating, the number of laser pulses in a certain time is limited and therefore a passive image can be obtained at no cost during laser recovery. The exposure profiles are designed to have the same passive component. 
The range-intensity profile $C(r)$ can be calibrated with measurements on targets with fixed albedo. We extract depth from three captures with different delays $\xi_i, i \in \{1,2,3\}$, resulting in a set of profiles $C_i(r)$ and measurements $I_i(r)$. We approximate the profiles with Chebychev polynomials of degree 6 as $\tilde{C}(r)$. Figure~\ref{fig:range_intensity_profile} shows the range-intensity profiles used in this work and their approximations, see supplemental material for details on the exposure profile design.
The final measurement, after read-out, is affected by photon shot noise and read-out noise as
\begin{align}
z = I(r)  + \eta_p \left( I(r) \right) + \eta_g ,
\label{eq:noise_model}
\end{align}
for a given pixel location, with $\eta_p$ being a Poissonian signal-dependent noise component and $\eta_g$ a Gaussian signal-independent component, which we adopt from~\cite{Foi2008}.

\vspace{-6pt}
\paragraph{Measurement Distortions}
A number of systematic and random measurement distortions make depth estimation from gated images challenging. Scene objects with low reflectance only return few signal photons, prohibiting an unambiguous mapping from intensities to depth and albedo in the presence of the Poissonian-Gaussian measurement fluctuations from Eq.~\eqref{eq:noise_model}. Systematic distortions include multi-path bounces of the flash illumination, see also~\cite{su2018deep}. In typical driving scenarios, severe multi-path reflection can occur due to wet roads acting as mirroring surfaces in the scene. Note that these are almost negligible in line or point-based scanning-lidar systems~\cite{achar2017epipolar}. Automotive applications require large laser sources that cannot be placed next to the camera, inevitably resulting in shadow regions without measurements available. Severe ambient sunlight, present as an offset in all slices, reduces the dynamic range of the gated measurements. In this work, we demonstrate a reconstruction architecture which addresses all of these issues in a data-driven approach, relying on readily available sparse lidar depth as training labels. Before describing the proposed approach, we introduce a per-pixel baseline estimation method.

\vspace{-6pt}
\paragraph{Per-Pixel Least-Squares Estimate.}
Ignoring all of the above measurement distortions, assuming no drift in the pulse and exposure profiles and Gaussian noise only in Eq.~\eqref{eq:noise_model}, an immediate baseline approach is the following per-pixel least-squares estimation. Specifically, for a single pixel, we stack the measurements ${z}_{\{1,2,3\}}$ for a sequence of delays ${\xi}_{\{1,2,3\}}$ in a single vector $\mathbf{z} = [ {z}_1, \ldots, {z}_3 ]$. We can estimate the depth and albedo jointly as
\begin{align}
\vspace{-4pt}
\hat{r}_{LS} = \minimize{r,\alpha} \left|\left| \mathbf{z} - \alpha \mathbf{\tilde{C}}(r) \right|\right|_2^2, \label{eq:ls}
\vspace{-4pt}
\end{align}
where $\mathbf{\tilde{C}}(r) = [ \tilde{C}_1(r), \ldots, \tilde{C}_3(r) ]$ is a Chebychev intensity profile vector. Since the range-intensity profiles are non-linear, we solve this nonlinear least-squares estimation using the Levenberg-Marquardt optimization method, see details in the supplemental document. 
\begin{figure}
	\centering
	\vspace{.1em}
	\resizebox{\columnwidth}{!}{
		\begin{tikzpicture}
				\begin{axis}
					[
					ylabel=$C_{i}(r)$,
					xmin=10, xmax=100,
					ymin=-100, ymax=1100,
					legend style={
						cells={anchor=west},
						legend pos=north east,
						font=\scriptsize
					},
					legend columns=3, 
					legend entries={$\tilde{C}_1(r)$, $\tilde{C}_2(r)$, $\tilde{C}_3(r)$},
					height=0.38\columnwidth,
					width=\linewidth
					]
				
				\addlegendimage{very thick, densely dashed, dai_deepred}
				\addlegendimage{very thick, densely dashed, dai_petrol}
				\addlegendimage{very thick, densely dashed, dai_ligth_grey40K}
				
				\addplot+ 	[
							thick,
							dai_ligth_grey40K,
							only marks,
							mark=x,
							mark options={color=dai_ligth_grey40K},
							mark size=2.5pt,
							]
				table[x index=0,y index=3,col sep=space]{fig/rip_real.txt};
				
				\addplot+ 	[
							thick,
							dai_petrol,
							only marks,
							mark=x,
							mark options={color=dai_petrol},
							mark size=2.5pt,
							]
				table[x index=0,y index=2,col sep=space]{fig/rip_real.txt};
				
				\addplot+ 	[
							thick,
							dai_deepred,
							only marks,
							mark=x,
							mark options={color=dai_deepred},
							mark size=2.5pt,
							]
				table[x index=0,y index=1,col sep=space]{fig/rip_real.txt};
				
				\addplot+ 	[
							mark=none,
							densely dashed,
							thick, 
							dai_deepred,
							]
				table[x index=0,y index=1, col sep=space]{fig/rip_cheb.txt};
								
				\addplot+ 	[
							mark=none,
							densely dashed,
							thick, 
							dai_petrol,
				]
				table[x index=0,y index=2, col sep=space]{fig/rip_cheb.txt};
								
				\addplot+ 	[
							mark=none,
							densely dashed,
							thick, 
							dai_ligth_grey40K,
							]
				table[x index=0,y index=3, col sep=space]{fig/rip_cheb.txt};
				
				\end{axis}
	\end{tikzpicture}
	}
	\vspace*{-20pt}
    \caption{Discrete measurements (marked with crosses) of the three range-intensity profiles $C_i(r), i \in \{1,2,3\}$ used in this work, and their continuous Chebychev approximations $\tilde{C}_i(r)$ plotted with distance $r$ [m].}
    \vspace{-1.5em}
    \label{fig:range_intensity_profile}
\end{figure}
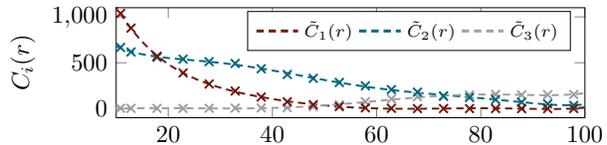
\begin{figure*}[t]
    \centering
    \includegraphics[width=\textwidth]{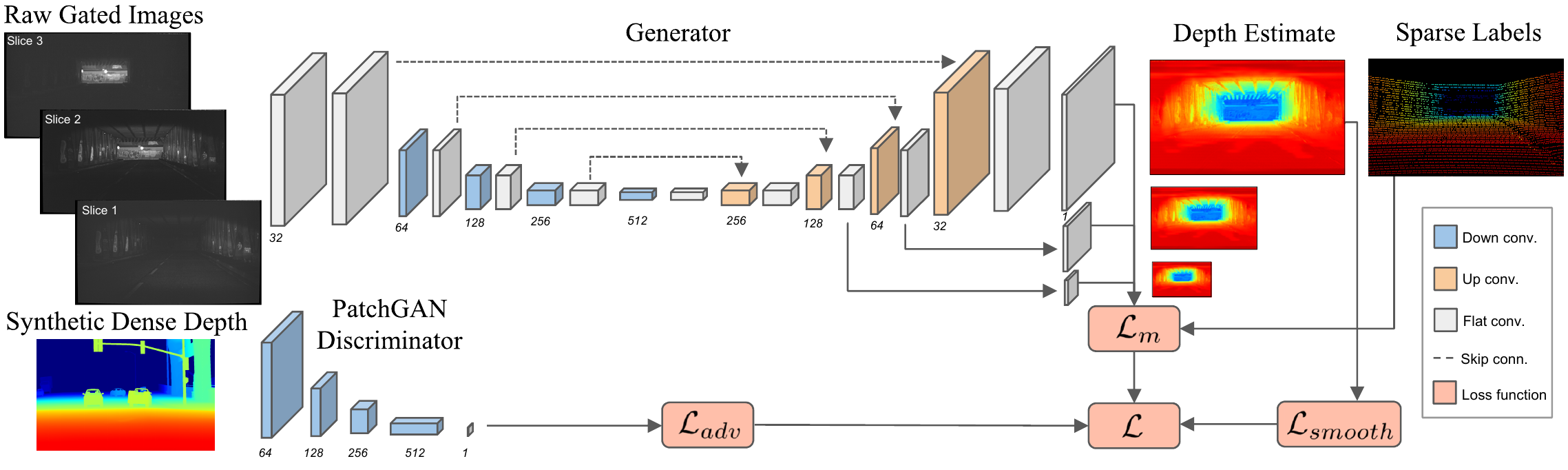}
    \vspace*{-20pt}
    \caption{The proposed \textsc{Gated2Depth} architecture estimates dense depth from a set of three gated images (actual reconstruction and real captures shown). To train the proposed generator network $G$ using sparse depth from lidar point samples, we rely on three loss-function components: a sparse multi-scale loss $\mathcal{L}_{\text{mult}}$ which penalizes sparse depth differences on three different binned scales, a smoothness loss $\mathcal{L}_{\text{smooth}}$, and an adversarial loss $\mathcal{L}_{\text{adv}}$. The adversarial loss incorporates a discriminator network which was trained on synthetic data, using a separate throw-away generator, and allows to transfer dense depth details from synthetic data without domain adaptation.}
    \label{fig:Networks}
    \vspace*{-10pt}
\end{figure*}

%------------------------------------------------------------------------
\section{Learning Depth from Gated Images} 
In this section, we introduce the \emph{Gated2DepthNet} network. 
The proposed model is the result of a systematic evaluation of different
input configurations, network architectures, and training schemes. 
We refer the readers to the supplemental document for 
a comprehensive study on all evaluated models.

The proposed network architecture is illustrated in Figure~\ref{fig:Networks}. The input to our network are three gated slices, allowing it
to exploit the corresponding semantics across the slices to estimate accurate pixel-wise depth. An immediately apparent issue for this architecture is that dense ground truth depth for large-scale scenes is not available. This issue becomes crucial when designing deep models that require large training datasets to avoid overfitting. We address this problem with a training strategy that transfers dense depth semantics learned on synthetic data to a network trained on sparse lidar data. 

The proposed \textit{Gated2DepthNet} is 
composed of a generator $G$, which we train for our 
dense depth estimation task. $G$ is a multi-scale variant 
of the popular U-net~\cite{ronneberger2015u} architecture. 
To transfer dense depth from synthetically generated depth 
maps to sensor data, we introduce a discriminator $D$, 
a variant of PatchGAN~\cite{pix2pix2016}, and train 
the network in a two-stage process. In the first stage, 
we train a network ($G$ and $D$) on synthetic data as 
generative adversarial network~\cite{Goodfellow2016}. The generator and discriminator 
are trained in alternating fashion in a least-square GAN~\cite{Mao:2017} approach: $G$ is trained to generate accurate dense depth estimations, \emph{using synthetic ground truth}, and to convince $D$ that the estimations correspond to a real depth maps;  $D$ is trained to detect whether a dense depth map comes from $G$ or is a real one. In the second stage, we train the network on real gated images that follow the target domain distribution. We now use sparse lidar measurements as groundtruth and keep the discriminator fixed. To use sparse lidar measurements in the final training stage, we introduce a multi-scale loss (see Section~\ref{sec:loss}) that penalizes differences to sparse lidar points by binning these to depth maps at multiple scales.

Our generator consists of 4 pairs of convolutions with a max pooling operation after each pair. The
encoder portion produces internal maps $\frac{1}{2}$, $\frac{1}{4}$, $\frac{1}{8}$, and $\frac{1}{16}$
of the original input size. The decoder consists of four
additional convolutions, and transposed convolutions after
each pair. As the depth estimate shares semantics with the input, we use symmetric skip connections, see Figure~\ref{fig:Networks}.

In the discriminator, we use a PatchGAN variant to best represent high-frequency image content. 
To this end, we define a fully convolutional network with five layers, each layer consisting of 
4x4 kernels with stride 2, and leaky  ReLUs with slope 0.2. The network classifies overlapping patches of a 
dense depth map instead of the whole map. 

\subsection{Loss Function}
\label{sec:loss}
We train our proposed network to minimize a three-component loss, $\mathcal{L}$, with each component modeling different statistics of the target depth
\begin{equation}
\mathcal{L} = \mathcal{L}_{\text{mult}} + \lambda_{s}  \mathcal{L}_{\text{smooth}} +
\lambda_{a}  \mathcal{L}_{\text{adv}}
\vspace{-9pt}
\end{equation}

\paragraph{Multi-scale loss ($\mathcal{L}_{\text{mult}}$)}
This loss component penalizes differences between the ground truth labels and the depth estimates. We define $\mathcal{L}_{\text{mult}}$ as a multi-scale loss over the generator's output $d$ and its corresponding target $\tilde{d}$
\begin{equation}
\mathcal{L}_{\text{mult}} ({d}, {\tilde{d}}) = \sum_{i=1}^{M}{\lambda_{m_i} \mathcal{L}_{\text{L1}}(d^{(i)}, \tilde{d}^{(i)})}, 
\vspace{-6pt}
\end{equation}
where $d^{(i)}$ and $\tilde{d}^{(i)}$ are the generator's output and target
at a scale $\small (i)$, $\mathcal{L}_{\text{L1}}(d^{(i)}, \tilde{d}^{(i)})$ is the
loss at scale $\small (i)$, and $\lambda_{m_i}$ is the weight of the
loss at the same scale. We define three scales ${1}/{2^i}$ with $i \in \{0,1,2\}$, binning as illustrated in Figure~\ref{fig:Networks}.
For a scale $\small (i)$, we define $L_{\text{L1}}(d^{(i)}, \tilde{d}^{(i)})$ as the mean absolute error
\begin{equation}
\vspace{-4pt}
\mathcal{L}_{\text{L1}}(d^{(i)}, \tilde{d}^{(i)}) = \frac{1}{N} \sum_{j, k}{|d^{(i)}_{jk} - \tilde{d}^{(i)}_{jk}|},
\vspace{-2pt}
\end{equation}
with the subscript $jk$ indicates here a discretized bin corresponding to pixel position $(j,k)$.
When training with synthetic data, 
we compute $\mathcal{L}_{\text{L1}}$ over all pixels.
For training with real data, we only compute this loss at bins that include at least one lidar sample point. $\mathcal{L}_{\text{L1}}$ is 
formally defined as
\begin{equation}
\vspace{-4pt}
\mathcal{L}_{\text{L1}}(d^{(i)}, \tilde{d}^{(i)}) = \frac{1}{N} \sum_{j, k}{|d^{(i)}_{jk} - \tilde{d}^{(i)}_{jk}| m^{(i)}_{jk}}
%\mathcal{L}_{L1} = \frac{1}{N} \sum_{i, j}{|d_{ij} - \tilde{d}_{ij}| m_{ij}}
\vspace{-2pt}
\end{equation}
where $m_{jk} = 1$ when the bin $(j,k)$ contains at least one lidar sample, and $m_{jk} = 0$ otherwise. 
For smaller scales, we average all samples per bin.

\paragraph{Weighted Smoothness Loss ($\mathcal{L}_{\text{smooth}}$)}
We rely on an additional smoothness loss $\mathcal{L}_{\text{smooth}}$ 
to regularize the depth estimates. Specifically we use a total variation loss weighted by the input image gradients~\cite{Ramirez2018}, that is
\begin{equation} \label{eq:tv}
\vspace{-4pt}
\hspace{-5pt}\mathcal{L}_{\text{smooth}} = \frac{1}{N} \sum_{i, j}{|\partial_x d_{i,j}|\epsilon ^{-|\partial_x z_{i,j}|} + |\partial_y d_{i, j}| \epsilon ^{-|\partial_y z_{i,j}|}},
\vspace{-2pt}
\end{equation}
where $z$ is here the input image.
As sparse lidar data is sampled on horizontal lines due to the rotating scanning setup, 
a generator trained on this data is biased to outputs with similar horizontal patterns. We found that increasing the weight of the 
vertical gradient relative to the horizontal one helps to mitigate  
this problem. 

\paragraph{Adversarial loss ($\mathcal{L}_{\text{adv}}$)}
We define the adversarial loss following~\cite{Mao:2017} with the PatchGAN~\cite{pix2pix2016} discriminator:
\begin{equation}
\vspace{-4pt}
\begin{aligned}
\mathcal{L}_{\text{adv}} = & \frac{1}{2} \mathbb{E}_{y\sim p_{\text{depth}}(y)} [(D(y) - 1)^2] + \\
 & \frac{1}{2} \mathbb{E}_{x\sim p_{\text{gated}}(x)} [(D(G(x)))^2]
\end{aligned}
\vspace{-2pt}
\end{equation}
Note the {discriminator is fixed in the second training stage}.

\subsection{Training and Implementation Details} \label{sec:training}
We use ADAM optimizer with the learning rate set to 0.0001.
For the global loss function, we experimentally determined $\lambda_{s} = 0.0001$ and $\lambda_{a} = 0.001$. 
For the multi-scale loss, we define $\lambda_{m_0} = 1$, $\lambda_{m_1} = 0.8$, and 
$\lambda_{m_2} = 0.6$. The full system runs at real-time rates of \unit[25]{Hz}, including all captures and inference (on a single TitanV).

%------------------------------------------------------------------------
\section{Datasets}
In this section, we describe the real and synthetic data sets used to train and evaluate the proposed method. 

\vspace{-9pt}
\paragraph{Real Dataset}
To the best of our knowledge, we provide the \emph{first long-range gated dataset}, covering snow, rain, urban and sub-urban driving during 4,000~km in-the-wild acquisition. To this end, we have equipped a testing vehicle with a standard RGB stereo camera (Aptina AR0230), lidar system (Velodyne HDL64-S3) and a gated camera (BrightwayVision BrightEye) with flood-light source integrated into the front bumper, shown in Figure~\ref{fig:Teaser}. Both cameras are mounted behind the windshield, while the lidar is mounted on the roof. The stereo camera runs at \unit[30]{Hz} with a resolution of 1920x1080 pixels. The gated camera provides \unit[10]{bit} images with a resolution of 1280x720 at a framerate of \unit[120]{Hz}, which we split up in three slices plus an additional ambient capture without active illumination. The car is equipped with two vertical-cavity surface-emitting laser (VCSEL) modules, which are diffused, with a wavelength of \unit[808]{nm} and a pulsed optical output peak power of \unit[500]{W} each. The peak power is limited due to eye-safety regulations. Our reference lidar systems is running with \unit[10]{Hz} and yields 64 lines. All sensors are calibrated and time-synchronized. During a four-week acquisition time in Germany, Denmark and Sweden, we recorded 17,686 frames in different cities (Hamburg, Kopenhagen, Gothenburg, V\r{a}rg\r{a}rda, Karlstad, \"Orebro, V\"aster\r{a}s, Stockholm, Uppsala, G\"avle, Sundsvall, Kiel). Figure~\ref{fig:dataset_distribution} visualizes the distribution of the full dataset, and Figure~\ref{fig:example_images_real} shows qualitative example measurements. We captured images during night and day and in various weather conditions (clear, snow, fog). The samples in clear weather conditions (14,277) are split into a training (7,478 day\,/\,4,460 night) and test set (1,789 day\,/\,550 night). Since snow and fog disturbs the lidar data, we do not use snowy nor foggy data for training.
\begin{figure}[t]
\vspace{-0.3cm}
\resizebox{.99\linewidth}{!}{
\begin{subfigure}[t]{\columnwidth}
	\begin{tikzpicture}
\begin{axis}[
	title=Real Dataset,
	symbolic x coords={clear, snow, fog},
	ylabel=Samples,
	xtick={clear, snow, fog},
	tick label style={/pgf/number format/fixed},
	scaled ticks=false,
	enlarge x limits=0.3,
	legend style={legend columns=-1},
	legend pos=north east,
	ybar=6pt,% configures `bar shift'
	bar width=20pt,
	width=0.98\columnwidth,
	height=0.6\columnwidth,
	ymin=0,
]
\addplot[fill=dai_deepred, dai_deepred] 
	coordinates {(clear,9267) (snow,1516) (fog,848)};

\addplot[fill=dai_petrol, dai_petrol]  
	coordinates {(clear,5010) (snow,788) (fog,257)};

\legend{Day,Night}
\end{axis}
\end{tikzpicture}
\end{subfigure}
\begin{subfigure}[t]{\columnwidth}
	\begin{tikzpicture}
\begin{axis}[
	title=Synthetic Dataset,
    symbolic x coords={urban, highway, overland},
	xtick={urban, highway, overland},
	enlarge x limits=0.3,
	legend style={legend columns=-1},
	legend pos=north east,
	ybar=6pt,
	bar width=20pt,
	width=0.98\columnwidth,
	height=0.6\columnwidth,
	ymin=0,
]
\addplot[fill=dai_deepred, dai_deepred] 
	coordinates {(urban,4470) (highway,1061) (overland,862)};

\addplot[fill=dai_petrol, dai_petrol] 
	coordinates {(urban,2201) (highway,593) (overland,617)};

\legend{Day,Night}
\end{axis}
\end{tikzpicture}
\end{subfigure}
}
\vspace{-15pt}
\caption{Dataset distribution.}
\label{fig:dataset_distribution}
\end{figure}
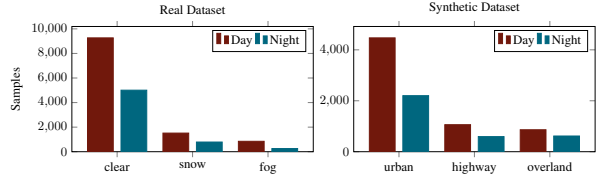
\begin{figure}[t!]
	\vspace{-0.3cm}
	\hspace{-3pt}
	\resizebox{.99\linewidth}{!}{
	\setlength\tabcolsep{1.5pt}
		\begin{tabular}{cccccc}
		
		\includegraphics[width=0.16\textwidth]{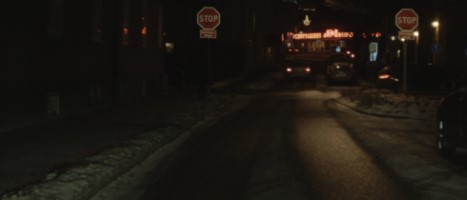} &
		\includegraphics[width=0.16\textwidth]{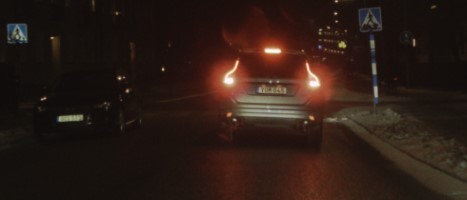} &
		\includegraphics[width=0.16\textwidth]{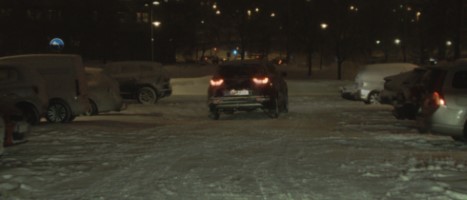} &
		\includegraphics[width=0.16\textwidth]{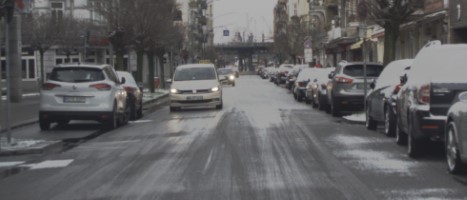} &
		\includegraphics[width=0.16\textwidth]{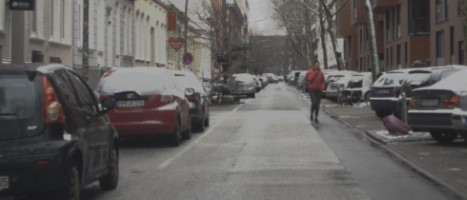} &
		\includegraphics[width=0.16\textwidth]{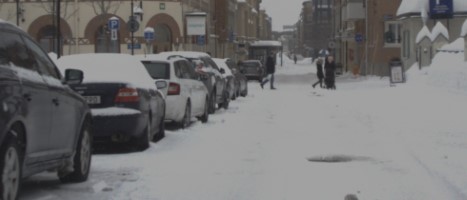} \\[2pt]
		
		\includegraphics[width=0.16\textwidth]{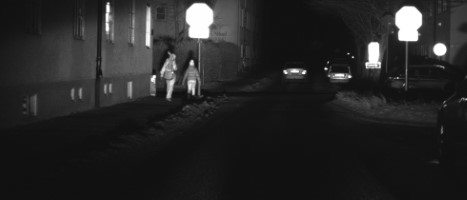} &
		\includegraphics[width=0.16\textwidth]{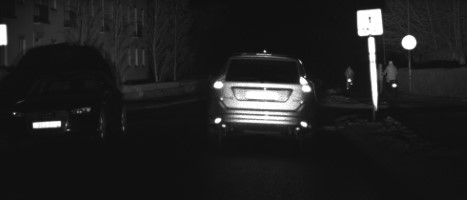} &
		\includegraphics[width=0.16\textwidth]{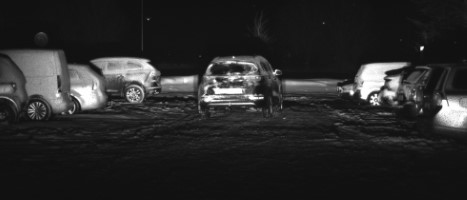} &
		\includegraphics[width=0.16\textwidth]{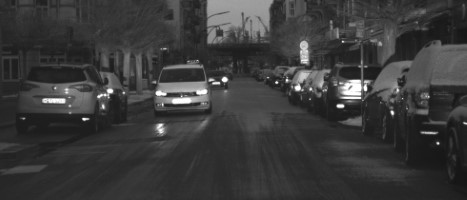} &
		\includegraphics[width=0.16\textwidth]{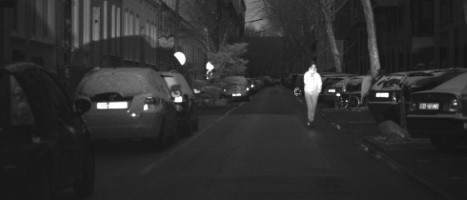} &
		\includegraphics[width=0.16\textwidth]{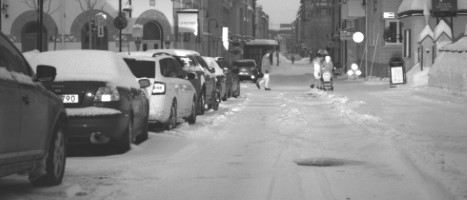} \\[2pt]
				
		\includegraphics[width=0.16\textwidth]{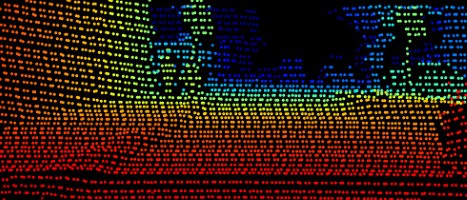} &
		\includegraphics[width=0.16\textwidth]{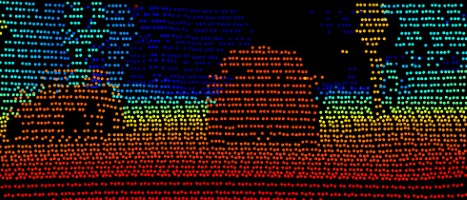} &
		\includegraphics[width=0.16\textwidth]{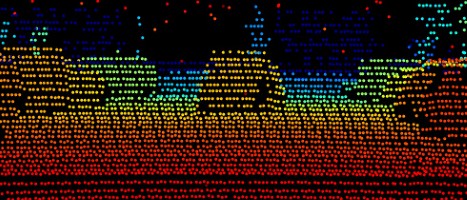} &
		\includegraphics[width=0.16\textwidth]{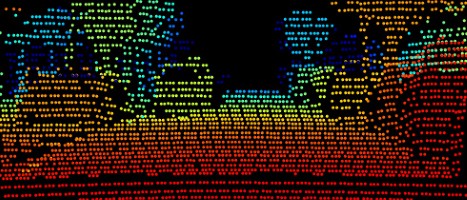} &
		\includegraphics[width=0.16\textwidth]{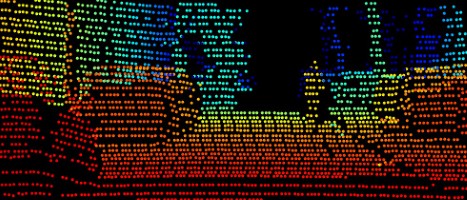} &
		\includegraphics[width=0.16\textwidth]{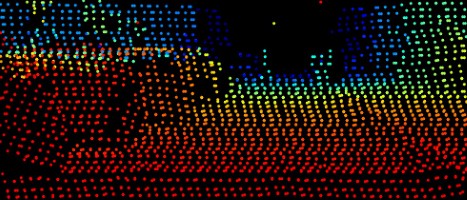} \\
					
		\end{tabular}	
	}
	\vspace{-0.3cm}
	\caption{Examples of real dataset (rgb/gated/lidar).}
	\label{fig:example_images_real}
\end{figure}
\begin{figure}[t!]
	\vspace{-0.3cm}
	\hspace{-3pt}
	\resizebox{.99\linewidth}{!}{
	\setlength\tabcolsep{1.5pt}
\begin{tabular}{cccccc}
		\includegraphics[width=0.16\textwidth]{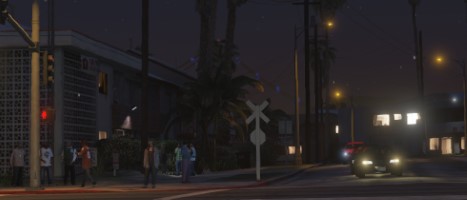} &
		\includegraphics[width=0.16\textwidth]{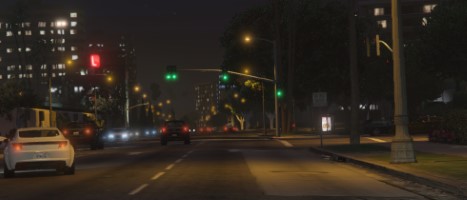} &
		\includegraphics[width=0.16\textwidth]{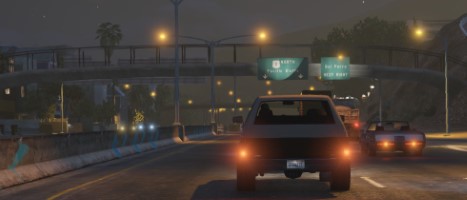} &
		\includegraphics[width=0.16\textwidth]{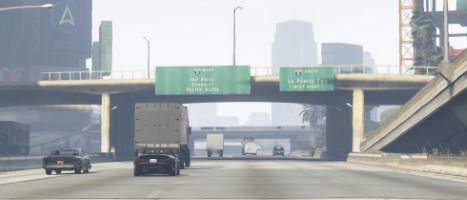} &
		\includegraphics[width=0.16\textwidth]{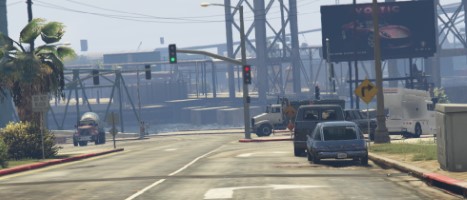} &
		\includegraphics[width=0.16\textwidth]{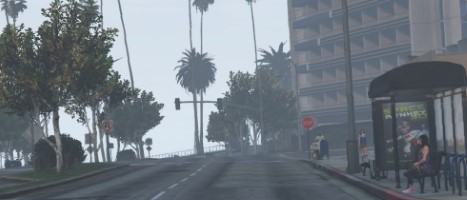} \\[2pt]
		
		\includegraphics[width=0.16\textwidth]{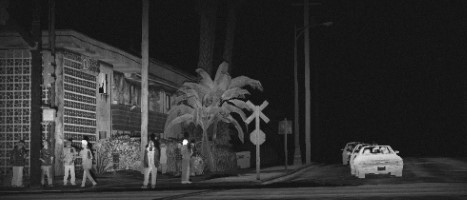} &
		\includegraphics[width=0.16\textwidth]{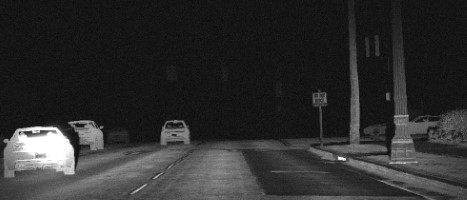} &
		\includegraphics[width=0.16\textwidth]{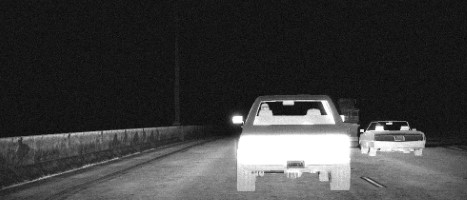} &
		\includegraphics[width=0.16\textwidth]{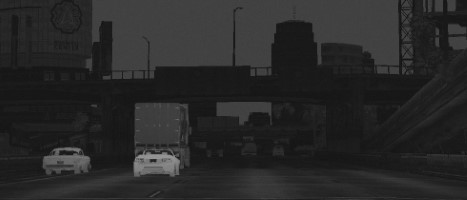} &
		\includegraphics[width=0.16\textwidth]{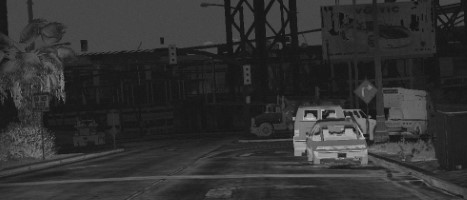} &
		\includegraphics[width=0.16\textwidth]{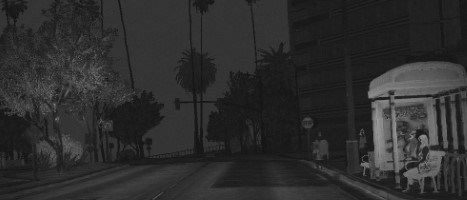} \\[2pt]
				
		\includegraphics[width=0.16\textwidth]{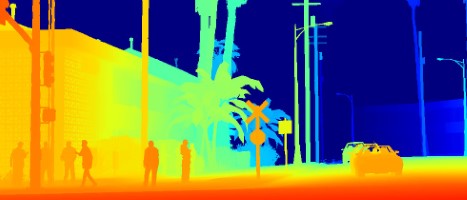} &
		\includegraphics[width=0.16\textwidth]{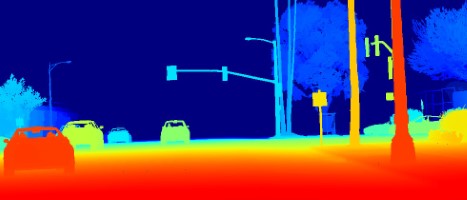} &
		\includegraphics[width=0.16\textwidth]{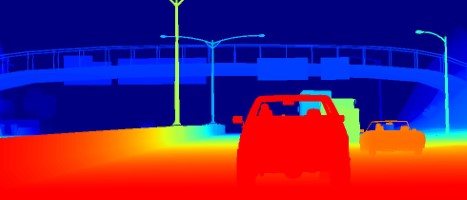} &
		\includegraphics[width=0.16\textwidth]{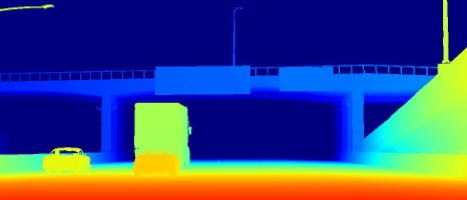} &
		\includegraphics[width=0.16\textwidth]{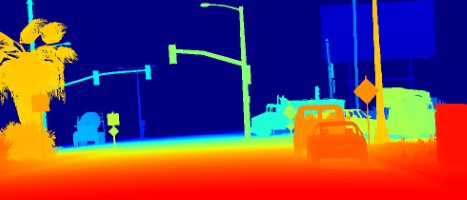} &
		\includegraphics[width=0.16\textwidth]{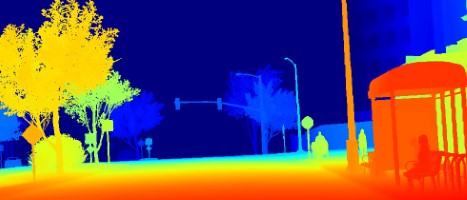} \\
			
		\end{tabular}
	}
	\vspace{-0.3cm}
	\caption{Examples of synthetic dataset (rgb/gated/depth). \vspace{-9pt}}
	\label{fig:example_images_synthetic}
\end{figure}

\vspace{-9pt}
\paragraph{Synthetic Dataset}\label{sec:synthetic_dataset}
While existing simulated datasets contain RGB and depth data, they do not provide enough information to synthesize realistic gated measurements that require {NIR} modeling and sunlight-illumination. We modify the GTA5-based simulator from~\cite{Richter2017} to address this issue. Please see the supplemental document for detailed description. We simulate 9,804 samples, and use 8,157 (5,279 day\,/\,2,878 night) for training and 1,647 (1,114 day\,/\,533 night) for testing. See Figure~\ref{fig:dataset_distribution} and Figure~\ref{fig:example_images_synthetic} for visualizations.

%------------------------------------------------------------------------
\section{Assessment} \label{sec:experiments}
\paragraph{Evaluation Setting}
We compare the proposed method against state-of-the-art depth estimation methods.
As per-pixel baseline methods, we compare to the least-squares baseline from Eq.~\eqref{eq:ls} and against the Bayesian estimate from Adam et al.~\cite{adam2017bayesian}. We compare against recent methods using monocular RGB images \cite{Godard2017},
stereo images \cite{Chang2018}, and RGB images in
combination with sparse lidar points \cite{ma2018sparse}.
For completeness, we also evaluate monocular depth estimation \cite{Godard2017}
applied on the integral of the gated slices, i.e. an actively illumination scene image without gating, which we dub full gated image.
Moreover, we also demonstrate Gated2Depth trained on full gated images only, validating the benefit of the coarse gating itself. 
For the method of Godard et al.~\cite{Godard2017}, we resized our images to the native size the model was trained on, as 
we noticed a substantial drop in performance when changing resolution at test time. For all other algorithms, we did not observe this behavior and we used the full resolution images.
For a fair comparison, we finetuned \cite{Godard2017} on RGB stereo pairs taken from the training set of our real dataset starting from the best available model.
For the comparisons in simulation, we calibrated the sampling pattern of the experimental lidar system and use this pattern for the Sparse-to-dense \cite{ma2018sparse} method. For \cite{Hirschmuller2008} we only had a hardware implementation available running in our test vehicle which does not allow synthetic evaluations.

We evaluate the methods with the metrics from \cite{eigen2014depth},
namely RMSE, MAE, ARD and $\delta_i < 1.25^i$ for $i \in \left\{1,2,3 \right\}$.
On the synthetic dataset, we compute the metrics over the whole depth maps.
On the real dataset, we compute the metrics only at the predicted pixels that
correspond to measured sparse lidar points. 
We observed that our lidar reference system degrades at distances 
larger than \unit[80]{m} and therefore we limit our evaluation to \unit[80]{m}.
For a fair comparison to methods that rely on laser illumination, 
we do not evaluate on non-illuminated pixels and introduce at 
the same time a completeness metric that describes on how many 
ground truth pixels is evaluated. Being $\left[ z_1, z_2, z_3 \right]$ a set of 
input gated slices, we define non-illuminated pixels as the ones that 
satisfy $\max ( \left[ z_1, z_2, z_3 \right] ) - \min ( \left[ z_1, z_2, z_3 \right] ) < 55$.
This definition allows us to avoid evaluation over outliers 
at extreme distances and with very low SNR.

\begin{table}[t]
    \footnotesize
    \setlength{\tabcolsep}{4pt} % general space between cols (6pt standard)
    \setlength\extrarowheight{2pt}
    \centering
    \resizebox{.99\linewidth}{!}{
    	\begin{tabular}{@{}lccccccc@{}}
            \toprule
            \multirow{2}{*}{\textbf{\textsc{Method}}}  & \textbf{RMSE}     & \textbf{ARD}   & \textbf{MAE}  & $\boldsymbol{\delta_1}$ & $\boldsymbol{\delta_2}$ & $\boldsymbol{\delta_3}$ & \textbf{Compl.} \\ 
			&  $\left[ m \right]$  &  & $\left[ m \right]$ & $\left[ \% \right]$ & $\left[ \% \right]$ & $\left[ \% \right]$ & $\left[ \% \right]$\\
			\midrule
			\multicolumn{8}{c}{\textbf{Simulated Data -- Night (Evaluated on Dense Ground Truth Depth)}} \\
			\midrule
            \textsc{Depth from Mono on RGB \cite{Godard2017}} & 74.40 & 0.62 & 58.47 & 7.76 & 13.67 & 29.17 & 100  \\ 
%& & \textsc{Depth from Mono on RGB \cite{Kuznietsov2017}} & 83.21 & 0.63 & 65.16 & 15.62 & 28.15 & 35.91 & 100  \\ 
\textsc{Depth from Mono on Full Gated \cite{Godard2017}} & 84.48 & 0.69 & 68.74 & 2.53 & 7.03 & 20.33 & 100  \\ 
%& & \textsc{Depth from Mono on Full Gated \cite{Kuznietsov2017}} & 90.44 & 0.70 & 73.03 & 9.19 & 18.23 & 24.77 & 100  \\ 
\textsc{Depth from Stereo \cite{Chang2018}} & 72.67 & 0.67 & 59.94 & 4.73 & 10.88 & 19.05 & 100  \\ 
%& & \textsc{Depth from Stereo \cite{pilzer2018unsupervised}} & 71.09 & 0.59 & 55.75 & 8.54 & 18.83 & 35.20 & 99  \\ 
\textsc{Sparse-to-Dense on Lidar (GT input) \cite{ma2018sparse}} & 64.08 & 0.33 & 42.33 & 56.74 & 63.19 & 67.87 & 100  \\ 
\textsc{Depth from ToF, Regression Tree} \cite{adam2017bayesian} & 40.33 & 0.45 & 26.03 & 37.33 & 55.96 & 68.47 & 45  \\ 
\textsc{Least Squares} & 30.45 & 0.29 & 18.66 & 60.82 & 77.41 & 83.61 & 34  \\ 
%\textsc{Gated2Depth} & \textbf{12.99} & \textbf{0.16} & \textbf{0.07} & \textbf{3.96} & \textbf{94.24} & \textbf{97.28} & \textbf{98.34} & 100  \\ 
\textsc{Gated2Depth} & 12.99 & 0.07 & 3.96 & 94.24 & 97.28 & 98.34 & 100  \\ 
			\midrule
			\multicolumn{8}{c}{\textbf{Simulated Data -- Day (Evaluated on Dense Ground Truth Depth)}} \\
			\midrule
			\textsc{Depth from Mono on RGB \cite{Godard2017}} & 75.68 & 0.63 & 59.95 & 6.27 & 14.14 & 28.28 & 100  \\ 
%& & \textsc{Depth from Mono on RGB \cite{Kuznietsov2017}} & 82.36 & 0.64 & 64.80 & 14.00 & 26.29 & 34.91 & 100  \\ 
\textsc{Depth from Mono on Full Gated \cite{Godard2017}} & 81.67 & 0.69 & 66.44 & 2.71 & 8.43 & 20.04 & 100  \\ 
%& & \textsc{Depth from Mono on Full Gated \cite{Kuznietsov2017}} & 85.50 & 0.69 & 68.75 & 9.14 & 17.86 & 24.73 & 100  \\ 
\textsc{Depth from Stereo \cite{Chang2018}} & 75.04 & 0.70 & 62.06 & 3.76 & 8.86 & 14.97 & 100  \\ 
%& & \textsc{Depth from Stereo \cite{pilzer2018unsupervised}} & 70.45 & 0.60 & 54.83 & 10.00 & 21.04 & 36.01 & 98  \\ 
\textsc{Sparse-to-Dense on Lidar (GT input) \cite{ma2018sparse}} & 60.97 & 0.31 & 39.63 & 58.84 & 65.30 & 69.77 & 100  \\ 
\textsc{Depth from ToF, Regression Tree} \cite{adam2017bayesian} & 27.17 & 0.52 & 20.05 & 25.53 & 47.77 & 66.30 & 23  \\ 
\textsc{Least Squares} & 15.52 & 0.36 & 10.32 & 55.44 & 73.29 & 82.35 & 16  \\ 
%\textsc{Gated2Depth} & \textbf{9.10} & \textbf{0.12} & \textbf{0.05} & \textbf{2.66} & \textbf{96.41} & \textbf{98.47} & \textbf{99.16} & 100  \\
\textsc{Gated2Depth} & 9.10 & 0.05 & 2.66 & 96.41 & 98.47 & 99.16 & 100  \\			    
			\midrule
			\multicolumn{8}{c}{\textbf{Real Data -- Night (Evaluated on Lidar Ground Truth Points)}} \\
			\midrule
			\textsc{Depth from Mono on RGB \cite{Godard2017}} & 16.87 & 0.38 & 11.64 & 21.74 & 63.15 & 80.96 & 100  \\ 
\textsc{Depth from Mono on RGB \cite{Godard2017} (FT)} & 11.41 & 0.23 & 6.18 & 76.64 & 89.53 & 94.19 & 100 \\
%& & \textsc{Depth from Mono on RGB \cite{Kuznietsov2017}} & 21.17 & 0.47 & 15.00 & 16.76 & 37.16 & 57.35 & 100  \\ 
\textsc{Depth from Mono on Full Gated \cite{Godard2017}} & 16.26 & 0.36 & 10.19 & 54.03 & 74.44 & 85.00 & 100  \\
\textsc{Depth from Mono on Full Gated \cite{Godard2017} (FT)} & 15.41 & 0.52 & 11.33 & 31.72 & 71.23 & 88.74 & 100  \\ 
%& & \textsc{Depth from Mono on Full Gated \cite{Kuznietsov2017}} & 23.01 & 0.52 & 17.04 & 10.16 & 24.80 & 43.68 & 100  \\ 
\textsc{Depth from Stereo \cite{Chang2018}} & 14.58 & 0.21 & 8.34 & 68.75 & 82.63 & 89.36 & 100  \\ 
%& & \textsc{Depth from Stereo \cite{pilzer2018unsupervised}} & 16.08 & 0.37 & 10.92 & 29.33 & 67.22 & 83.60 & 100  \\ 
\textsc{Depth from Stereo \cite{Hirschmuller2008}} & 15.51 & 0.36 & 8.75 & 63.94 & 76.19 & 82.31 & 63  \\ 
\textsc{Sparse-to-Dense on Lidar (GT input) \cite{ma2018sparse}} & 8.79 & 0.21 & 4.38 & 87.64 & 93.74 & 95.88 & 100  \\ 
\textsc{Depth from ToF, Regression Tree} \cite{adam2017bayesian} & 10.54 & 0.24 & 6.01 & 76.73 & 89.74 & 93.45 & 40  \\ 
\textsc{Least Squares} & 13.13 & 0.42 & 8.88 & 43.60 & 55.80 & 63.54 & 31  \\ 
%\textsc{Gated2Depth} & \textbf{8.39} & \textbf{0.26} & \textbf{0.15} & \textbf{3.79} & \textbf{87.52} & \textbf{93.00} & \textbf{95.21} & 100  \\ 
\textsc{Gated2Depth - Full Gated} & 14.86 & 0.29 & 8.84 & 58.79 & 58.79 & 79.84& 100 \\
\textsc{Gated2Depth} & 8.39 & 0.15 & 3.79 & 87.52 & 93.00 & 95.21 & 100  \\ 
			\midrule
			\multicolumn{8}{c}{\textbf{Real Data -- Day (Evaluated on Lidar Ground Truth Points)}} \\
			\midrule
			\textsc{Depth from Mono on RGB \cite{Godard2017}} & 17.67 & 0.37 & 12.28 & 13.87 & 60.93 & 79.17 & 100  \\ 
\textsc{Depth from Mono on RGB \cite{Godard2017} (FT)} & 10.24 & 0.18 & 5.47 & 80.49 & 91.78 & 95.61 & 100 \\
%& & \textsc{Depth from Mono on RGB \cite{Kuznietsov2017}} & 21.05 & 0.47 & 15.21 & 13.42 & 30.90 & 52.60 & 100  \\ 
\textsc{Depth from Mono on Full Gated \cite{Godard2017}} & 13.89 & 0.24 & 8.50 & 60.05 & 79.62 & 89.92 & 100  \\ 
\textsc{Depth from Mono on Full Gated \cite{Godard2017} (FT)} & 13.33 & 0.40 & 9.51 & 36.64 & 81.63 & 92.86 & 100 \\
%& & \textsc{Depth from Mono on Full Gated \cite{Kuznietsov2017}} & 21.49 & 0.47 & 15.75 & 12.89 & 30.20 & 51.22 & 100  \\ 
\textsc{Depth from Stereo \cite{Chang2018}} & 13.94 & 0.19 & 7.78 & 71.32 & 84.67 & 91.38 & 100  \\ 
%& & \textsc{Depth from Stereo \cite{pilzer2018unsupervised}} & 17.18 & 0.34 & 11.56 & 27.76 & 64.59 & 81.10 & 100  \\ 
\textsc{Depth from Stereo \cite{Hirschmuller2008}} & 9.63 & 0.17 & 4.59 & 85.80 & 92.72 & 95.20 & 86  \\ 
\textsc{Sparse-to-Dense on Lidar (GT input) \cite{ma2018sparse}} & 8.21 & 0.16 & 4.05 & 88.52 & 94.71 & 96.87 & 100  \\ 
\textsc{Depth from ToF, Regression Tree} \cite{adam2017bayesian} & 15.83 & 0.49 & 11.40 & 56.30 & 75.54 & 82.45 & 23  \\ 
\textsc{Least Squares} & 19.52 & 0.75 & 14.05 & 43.42 & 54.63 & 63.76 & 16  \\ 
%\textsc{Gated2Depth} & \textbf{7.61} & \textbf{0.22} & \textbf{0.12} & \textbf{3.53} & \textbf{88.07} & \textbf{94.32} & \textbf{96.60} & 100  \\
\textsc{Gated2Depth - Full Gated} & 13.75 & 0.26 & 8.16 & 62.48 & 62.48 & 82.93& 100 \\
\textsc{Gated2Depth} & 7.61 & 0.12 & 3.53 & 88.07 & 94.32 & 96.60 & 100  \\
 
            \bottomrule
        \end{tabular}
    }
    \vspace*{-3pt}
    \caption{\label{tab:results} Comparison of our proposed framework and
	state-of-the-art methods on unseen synthetic and real test data sets. 
	\textsc{GT input}: uses sparse ground truth as input. \textsc{FT}: model finetuned on our real data.
	\vspace*{-12pt}
}
\end{table}
\setlength\tabcolsep{1.5pt}
\begin{figure*}[t!]
\vspace{-0.5cm}
\hspace{-3pt}\begin{subfigure}{1.04\linewidth}
	\resizebox{.99\linewidth}{!}{
	\small
	{\scriptsize
\begin{tabular}{ccccccc}
	RGB & 
	Full Gated & 
	Lidar &  
	Gated2Depth & 
	Gated2Depth - Full Gated &
	Least-Squares &
	[m] \\
	
	\includegraphics[width=0.14285714285714285\textwidth]{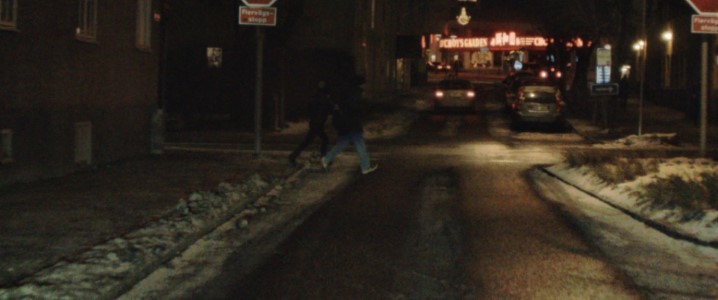} &
	\includegraphics[width=0.14285714285714285\textwidth]{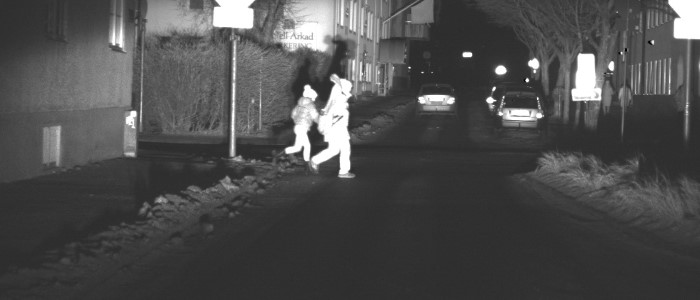} &
	\includegraphics[width=0.14285714285714285\textwidth]{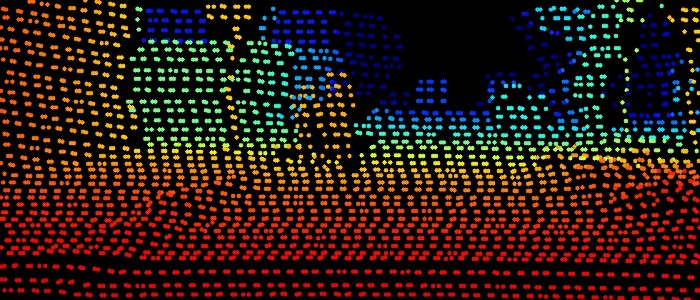} &
	\includegraphics[width=0.14285714285714285\textwidth]{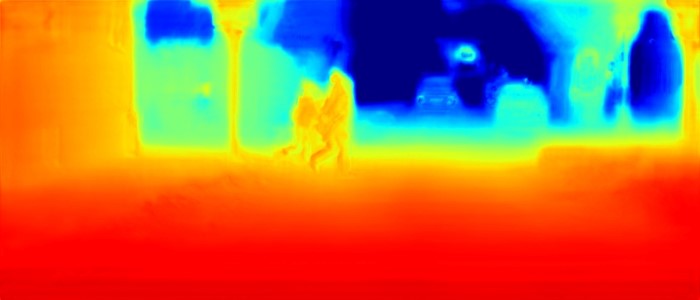} &
	\includegraphics[width=0.14285714285714285\textwidth]{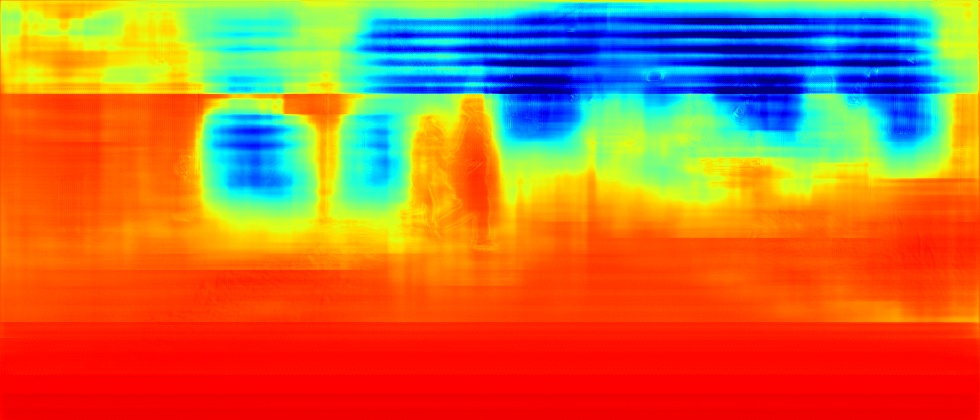} &
	\includegraphics[width=0.14285714285714285\textwidth]{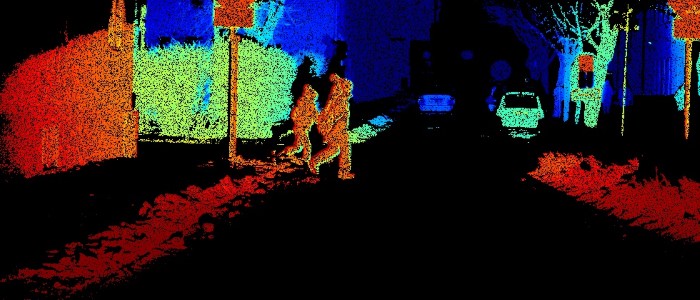} &
	\multirow{3}{*}[1.5cm]{\vspace{-3.03cm}\includegraphics[height=0.14\textwidth]{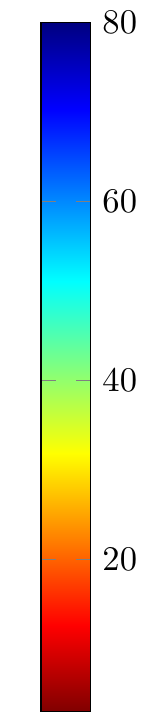}}
	\\
	
	Regression Tree \cite{adam2017bayesian} &
	Lidar+RGB  \cite{ma2018sparse} &
	Stereo \cite{Hirschmuller2008} &
	Stereo \cite{Chang2018} & 
	Mono Gated \cite{Godard2017} (FT) & 
	Monocular \cite{Godard2017} (FT) &
	\\
	
	\includegraphics[width=0.14285714285714285\textwidth]{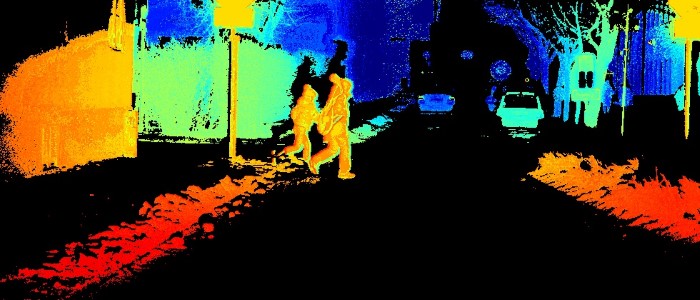} &
	\includegraphics[width=0.14285714285714285\textwidth]{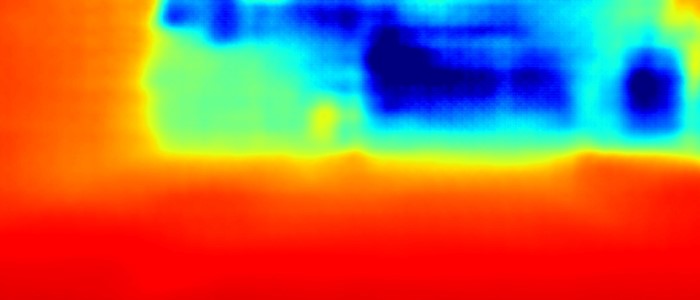} &
	\includegraphics[width=0.14285714285714285\textwidth]{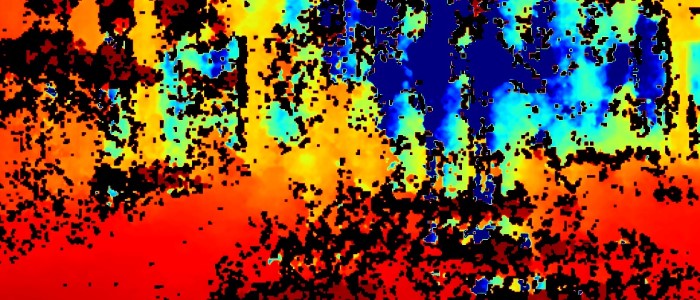} &
	\includegraphics[width=0.14285714285714285\textwidth]{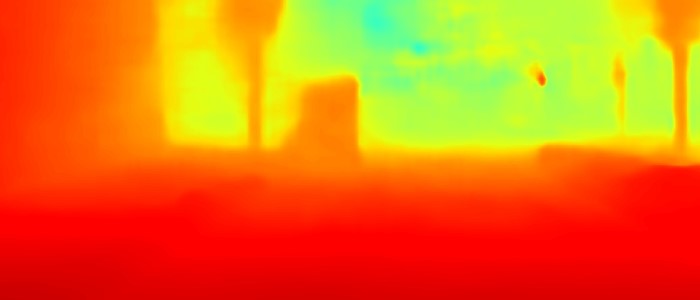} &
	\includegraphics[width=0.14285714285714285\textwidth]{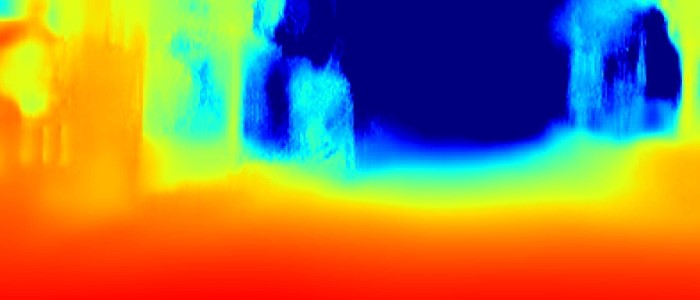} &
	\includegraphics[width=0.14285714285714285\textwidth]{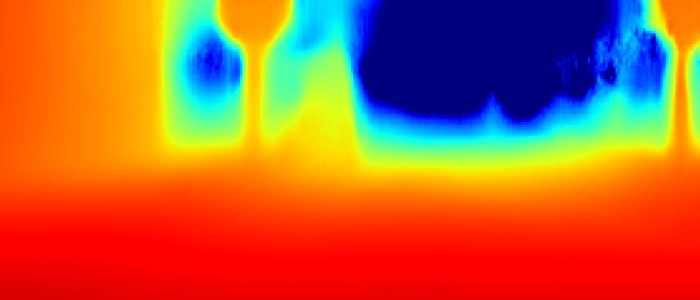} &
    \\
	
\end{tabular}
}
	}
	\vspace*{-5pt}
	\subcaption{Experimental night time results.}
	\label{fig:qualitative_results_real_night}
	\vspace*{5pt}
\end{subfigure}
    
\hspace{-3pt}\begin{subfigure}{1.03\linewidth} %HACK
    \resizebox{\linewidth}{!}{
    \small
	{\scriptsize
\begin{tabular}{ccccccc}
	RGB & 
	Full Gated & 
	Lidar &  
	Gated2Depth & 
	Gated2Depth - Full Gated &
	Least-Squares &
	[m]
	\\
	
	\includegraphics[width=0.14285714285714285\textwidth]{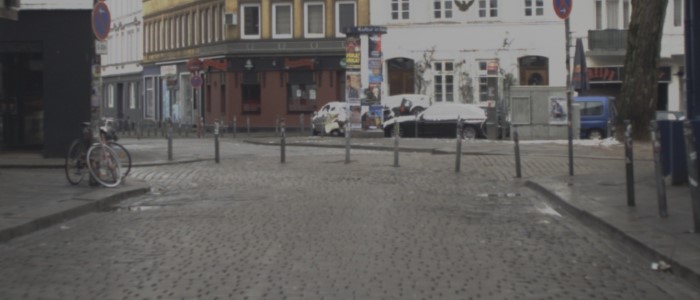} &
	\includegraphics[width=0.14285714285714285\textwidth]{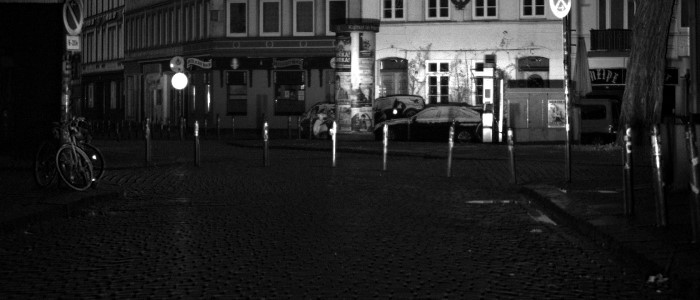} &
	\includegraphics[width=0.14285714285714285\textwidth]{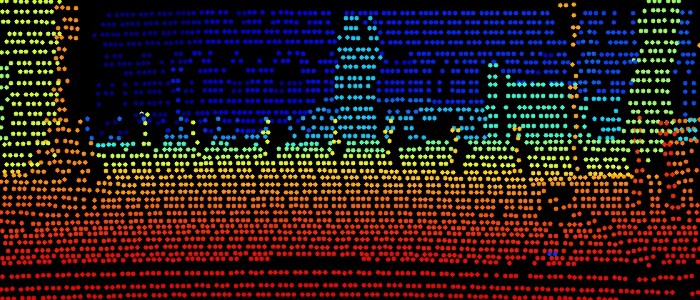} &
	\includegraphics[width=0.14285714285714285\textwidth]{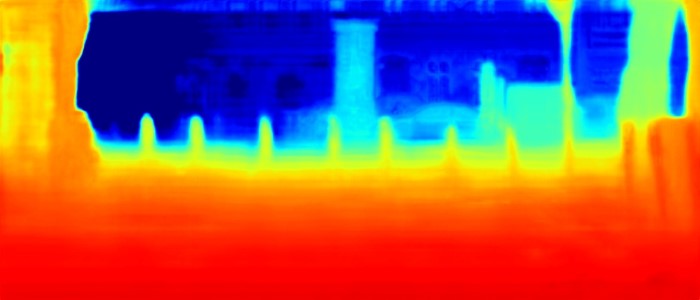} &
	\includegraphics[width=0.14285714285714285\textwidth]{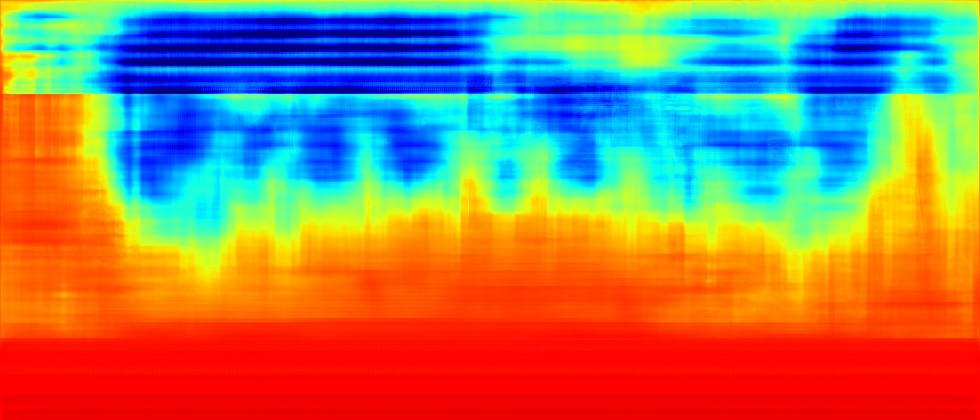} & 
	\includegraphics[width=0.14285714285714285\textwidth]{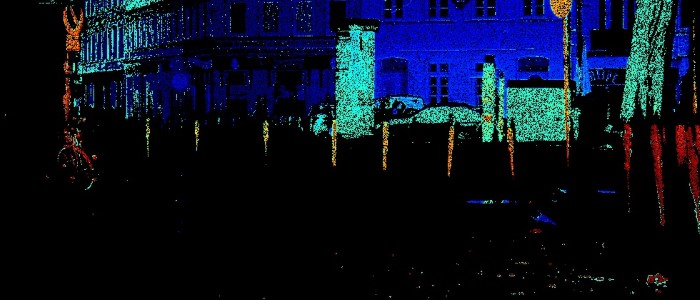} &
	\multirow{3}{*}[1.5cm]{\vspace{-3.03cm}\includegraphics[height=0.14\textwidth]{fig/colorbars/colorbar_vertical_real.pdf}}
	\\
	
	Regression Tree \cite{adam2017bayesian} &
	Lidar+RGB  \cite{ma2018sparse} &
	Stereo \cite{Hirschmuller2008} &
	Stereo \cite{Chang2018} & 
	Mono Gated \cite{Godard2017} (FT) & 
	Monocular \cite{Godard2017} (FT)
	\\
	
	\includegraphics[width=0.14285714285714285\textwidth]{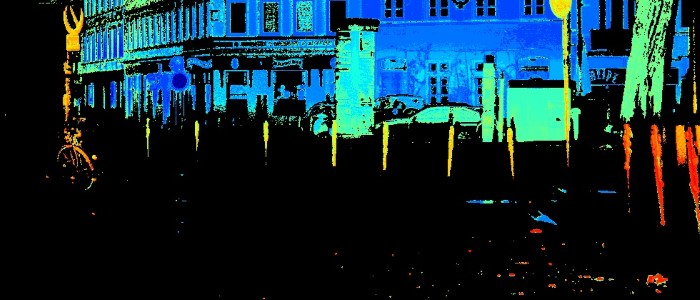} &
	\includegraphics[width=0.14285714285714285\textwidth]{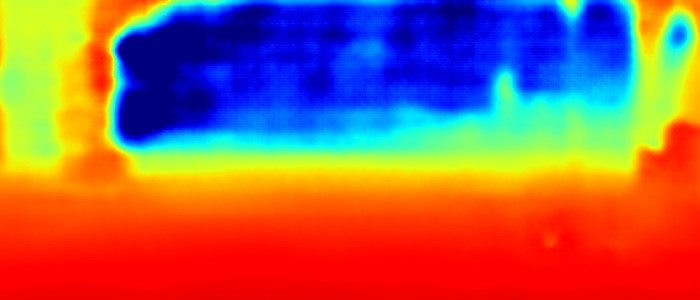} &
	\includegraphics[width=0.14285714285714285\textwidth]{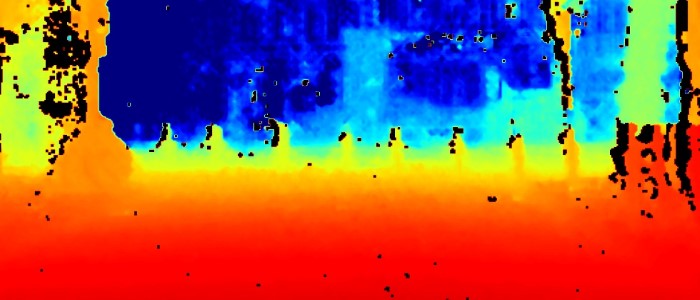} &
	\includegraphics[width=0.14285714285714285\textwidth]{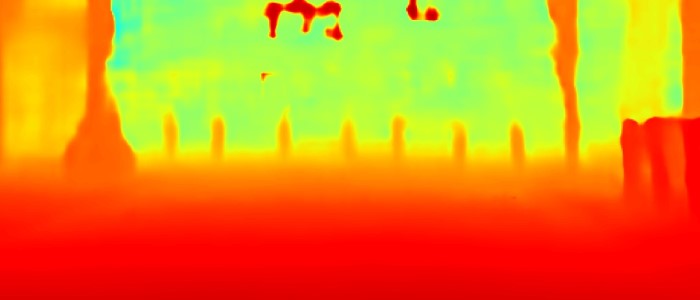} &
	\includegraphics[width=0.14285714285714285\textwidth]{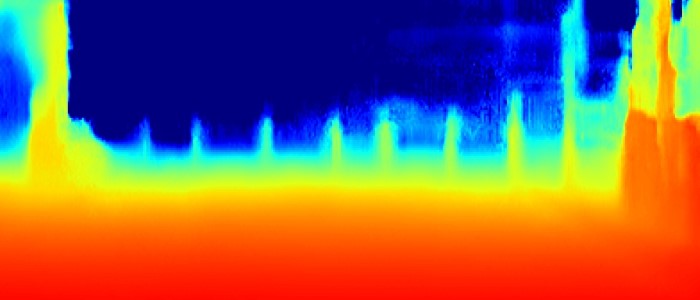} &
	\includegraphics[width=0.14285714285714285\textwidth]{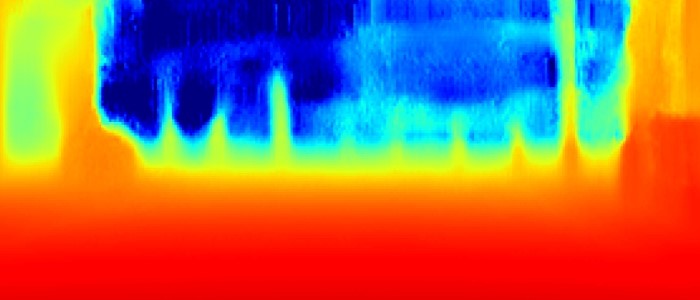}
	 \\
\end{tabular}
}
    }
    \vspace*{-5pt}
    \subcaption{Experimental day time results.}
    \label{fig:qualitative_results_real_day}
    \vspace*{5pt}
\end{subfigure}

\hspace{-3pt}\begin{subfigure}{1.04\linewidth}
	\resizebox{.99\linewidth}{!}{
	\small
	{\scriptsize
\begin{tabular}{ccccccc}
	RGB & 
	Full Gated &
	Depth GT & 
	Gated2Depth & 
	Least-Squares & 
	Regression Tree & 
	\hspace{-1mm}[m] \\
	
	\includegraphics[width=0.14285714285714285\textwidth]{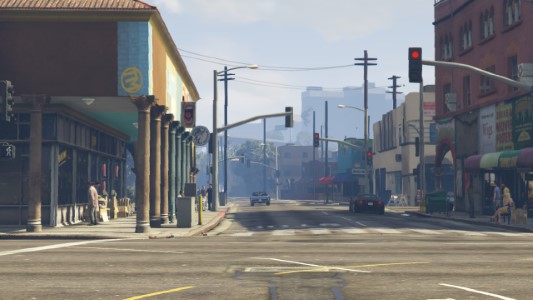} &
	\includegraphics[width=0.14285714285714285\textwidth]{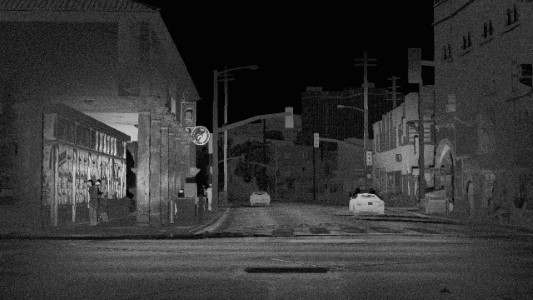} &
	\includegraphics[width=0.14285714285714285\textwidth]{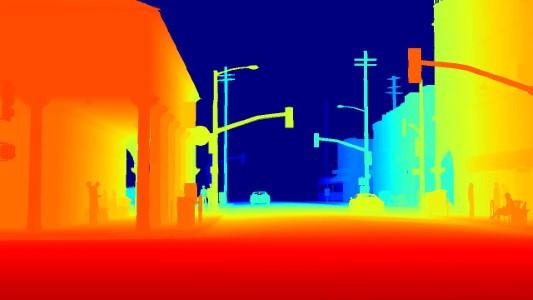} &
	\includegraphics[width=0.14285714285714285\textwidth]{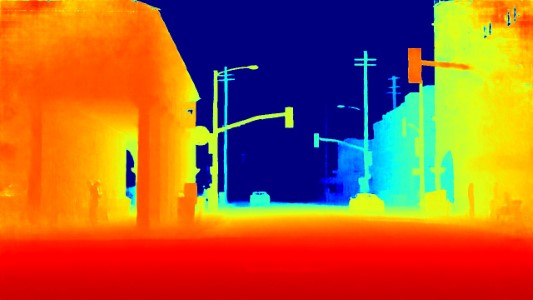} &
	\includegraphics[width=0.14285714285714285\textwidth]{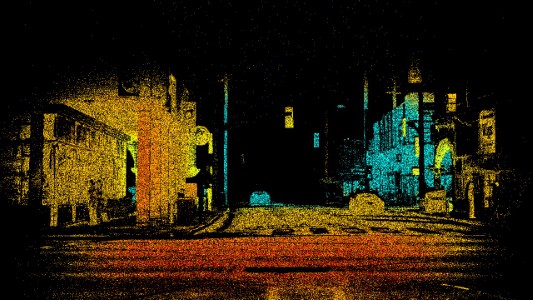} &
	\includegraphics[width=0.14285714285714285\textwidth]{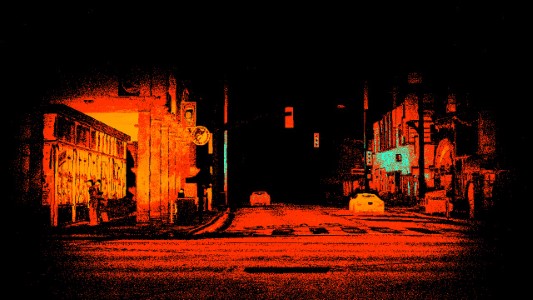} &
	\includegraphics[height=0.08\textwidth]{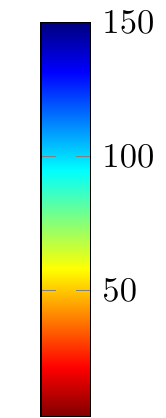}
	\\	
\end{tabular}
}
	}
	\vspace*{-3pt}
	\subcaption{Daytime simulation results.}
	\label{fig:qualitative_results_simulated_day}
	\vspace*{-7pt}
\end{subfigure}
	\caption{Qualitative results for our method and reference methods over real and
	synthetic examples. For each example, we include the corresponding RGB and full gated image,
	along with the lidar measurements. Our method generates more accurate and detailed maps over different distance
	ranges of the scenes in comparison to the other methods. For the simulation results in (c) we only show models finetuned on simulated data.}
	\label{fig:qualitative_results}
\vspace*{-6pt}
\end{figure*}
\subsection{Results on Synthetic Dataset}
Table~\ref{tab:results} (top) shows that the proposed method outperforms all other
reference methods by a large margin. The second-best method without gated images is the
depth completion based on lidar and RGB \cite{ma2018self}, which yields better results than monocular or stereo methods because 
it uses sparse lidar ground truth samples as input.
While monocular approaches struggle to recover absolute scale,
stereo methods achieve low accuracy over the large distance range due to the limited baseline.

Figure~\ref{fig:qualitative_results_simulated_day} shows 
an output example of our method and compares it with others. 
Our method captures better fine-grained details of a scene at 
both close and far distances.

\subsection{Results on Real Dataset}
Table~\ref{tab:results} (bottom) shows that the proposed method outperforms all compared methods, 
including the one that uses ground truth lidar points as input~\cite{ma2018sparse}. 
Hence, the method achieves high depth accuracy comparable to scanning lidar systems, while, in contrast, 
providing dense depth.
Moreover, Table~\ref{tab:results} validates the benefit of using multiple slices 
compared to a single continuously illuminated image.

\begin{figure}[t]
\vspace{-0.1em}
    \centering 
    \footnotesize
\setlength\tabcolsep{1pt}
\begin{tabular}{ccc}
    RGB & Lidar in Snow & Gated2Depth \\
	\includegraphics[height=0.137\linewidth]{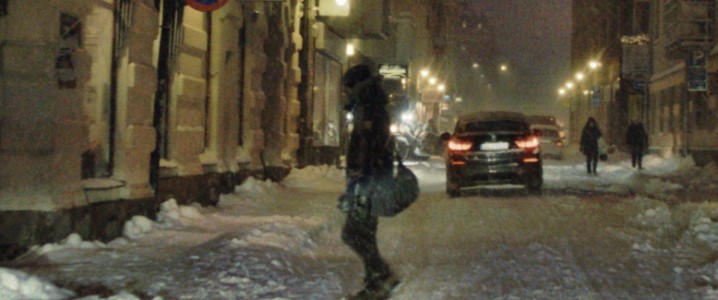} &
	\includegraphics[height=0.137\linewidth]{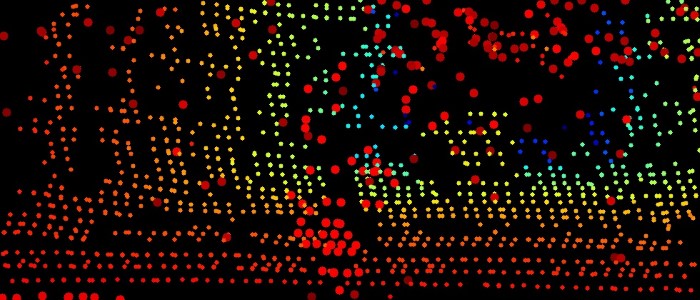} & 
	\includegraphics[height=0.137\linewidth]{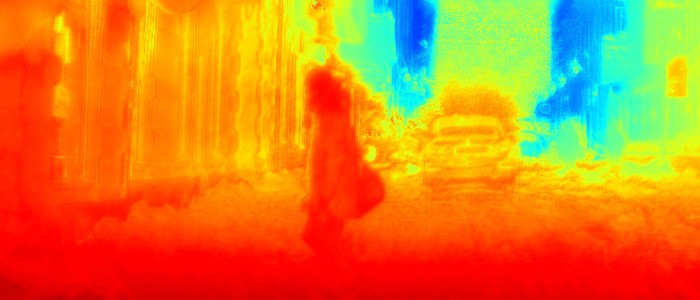}
\end{tabular} 
    \vspace*{-5pt}
    \caption{Results for strong backscatter in snow, with lidar clutter (larger points) around a pedestrian and in the sky.}
    \vspace*{-5pt}
    \label{fig:depth_backscatter}
\end{figure}

Figures~\ref{fig:qualitative_results_real_night} and~\ref{fig:qualitative_results_real_day} visualizes the dense depth estimation, and scene details captured by our method in comparison to state-of-the-art methods. 
Especially for fine details around pedestrians or small scene objects, the proposed method achieves higher resolution. 
In the example from Figure~\ref{fig:qualitative_results_real_night} our method shows all scene objects (two pedestrians, two cars), which are also recovered in both gated per-pixel estimation methods, but not at high density. While the sparse depth completion method misses major scene objects, our method preserves all of them. The same can be observed in the second example for the posts and the advertising column in Figure~\ref{fig:qualitative_results_real_day}.
Figure~\ref{fig:depth_backscatter} illustrates the robustness of our method in (unseen) snowing conditions. 
While the lidar shows strong clutter, our method provides a very clear depth estimation, as a by-product of the gated imaging acquisition itself.
\begin{figure}[t]
\vspace{-0.1em}
    \centering
    \footnotesize
\setlength\tabcolsep{1pt}
\begin{tabular}{ccc}	
	RGB & Least-Squares & Gated2Depth \\
	\includegraphics[height=0.137\linewidth]{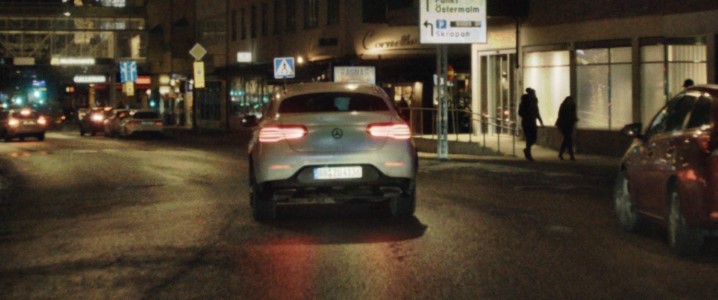} &
	\includegraphics[height=0.137\linewidth]{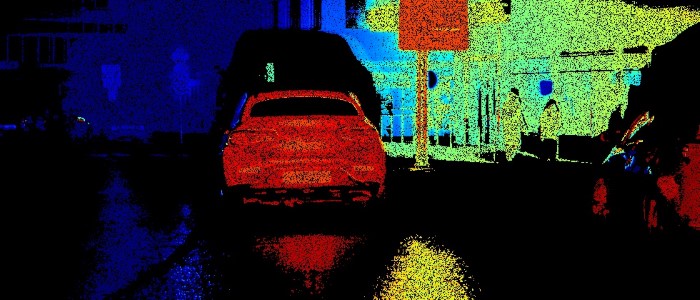} &
	\includegraphics[height=0.137\linewidth]{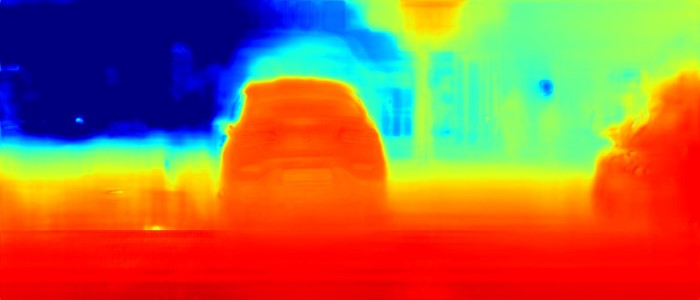} \\
\end{tabular}
	\vspace*{-5pt}
	\caption{Multipath Interference. In contrast to existing methods, such as the Least-Squares method, our method eliminates most multi-path interference (on the road here).}
	\label{fig:qualitative_results_multipath}
	\vspace*{-10pt}
\end{figure}
Figure~\ref{fig:qualitative_results_multipath} compares per-pixel estimation with the proposed approach.
The proposed method is able to fill in shadows and surfaces with low reflectance. Multi-path interference is suppressed by using the contextual information present in the whole image.

%------------------------------------------------------------------------
\section{Conclusions and Future Work}
In this work, we turn a CMOS gated camera into a cost-sensitive high-resolution dense flash lidar. 
We propose a novel way of transfer learning that allows us to leverage datasets 
with sparse depth labels for dense depth estimation. The proposed method outperforms 
state-of-the-art methods, which we validate in simulation and experimentally on outdoor 
captures with large depth range of up to \unit[80]{m} (limited by the range of the scanned reference lidar system).

An interesting direction for future research is the inclusion of RGB data, which could provide additional depth clues in areas with little variational information in the gated images. However, fusing RGB images naively as an additional input channel to the proposed architecture would lead to severe bias for distortions due to backscatter, see Figure~\ref{fig:sensor_performance}, which is properly handled by the proposed system. Exciting future applications of the proposed method include large-scale semantic scene understanding and action recognition using the proposed architecture either for dataset generation or in an end-to-end-fashion.

%------------------------------------------------------------------------

\vspace*{5mm}
\noindent\small{

This work has received funding from the European Union under the H2020 ECSEL Programme as part of the DENSE project, contract number 692449. Werner Ritter supervised this project at Daimler AG, and Klaus Dietmayer supervised the project portion at Ulm University. We thank Robert Böhler, Stefanie Walz and Yao Wang for help processing the large dataset. We thank Fahim Mannan for fruitful discussions and comments on the manuscript.}

%------------------------------------------------------------------------
\clearpage
{\small
\bibliographystyle{ieee_fullname}
\bibliography{bib}
}

\end{document}